\documentclass[pmlr]{jmlr}

\usepackage{graphicx}
\usepackage{booktabs}
\usepackage{longtable}
\usepackage{multirow}
\usepackage{xcolor}
\usepackage{colortbl}
\makeatletter
\def\set@curr@file#1{\def\@curr@file{#1}}
\makeatother
\usepackage{float}
\usepackage{amssymb}
\usepackage{placeins}

\newcommand{\bcr}{\textsc{BCR}}
\newcommand{\eca}{\textsc{ECA}}
\newcommand{\medec}{\textsc{MEDEC}}

\jmlrproceedings{Preprint}{Preprint}
\jmlrpages{}
\jmlryear{2026}
\jmlrworkshop{Accepted at MLHC 2026}

\makeatletter
\gdef\@reprint{Author version, August 2026.}
\makeatother

\title{Toward Better Assessment of LLMs' Performance in Clinical Error Detection}

\author{\Name{Yifan Zhang$^{1}$} \Email{eyfzh@udel.edu}\\
        \Name{Rahmatollah Beheshti$^{1}$} \Email{rbi@udel.edu}\\
        \addr $^{1}$ University of Delaware, Newark, DE, USA}

\begin{document}

\maketitle

\begin{abstract}
Automated detection of errors in clinical documentation is a promising application of large language models (LLMs), yet decisions to deploy such models rest on benchmarks that evaluate each clinical note in isolation. Error-detection benchmarks are typically constructed by injecting errors into notes, such that each erroneous note has a natural counterpart. Aggregate discriminative metrics (e.g., balanced accuracy or F1) do not exploit this structure. We show that this omission is consequential. In particular, evaluating 15 diverse LLMs on 4 standardized clinical error-detection test sets across 3 languages, we find that 13 of 15 models fall below the level of random pairwise discrimination, even while achieving F1 scores that standard practice would read as moderate. We also observe that the underlying bias patterns differ across languages: the same model can default to ``no error'' on one language and over-flag errors on another. To diagnose where discrimination breaks down, we further introduce a procedure to score the evidence models cite in their outputs. We find that while models consistently locate error-relevant content, they fail to produce the corresponding correct verdict on the clean counterpart. Finally, we show that F1 and pairwise accuracy are driven in opposite directions by the same underlying bias, so that ranking models by F1 may systematically promote the weakest discriminators. For safety-critical clinical NLP applications, we advocate for supplementing aggregate metrics with paired evaluations in benchmark reporting. Code and analysis scripts are available at \url{https://github.com/healthylaife/paired-clinical-eval}.
\end{abstract}

\section{Introduction}
\label{sec:intro}

Medical errors are a substantial and persistent source of preventable harm in US healthcare, with widely cited estimates ranging from tens of thousands to several hundred thousand deaths per year~\citep{iom1999err, makary2016medical}. A substantial portion of these errors originates in or propagates through clinical documentation: incorrect diagnoses recorded in progress notes, inappropriate treatment plans, or misidentified pathogens that cascade into downstream care decisions~\citep{BenAbacha2024_MEDIQA-CORR}. Manual review of clinical notes is time-consuming, inconsistent, and does not scale with growing documentation volumes. Automated detection using language models to flag potential errors is therefore of direct clinical interest.

From a machine learning perspective, clinical error detection is a particularly demanding task. Unlike medical question answering, where a model selects among predefined options, error detection requires the model to evaluate clinical reasoning. The model must read a note, understand the implicit chain from symptoms through diagnosis to management, and judge whether that reasoning is sound. This demands not only medical knowledge but also the ability to distinguish incorrect clinical decisions from decisions that are unconventional but clinically appropriate.

Recent work has established benchmarks for this task (\medec{}~\citep{BenAbacha2025_MEDEC}, MedRECT~\citep{Iwase2025_MedRECT}, and MedErrBench~\citep{ma2026mederrbench}), with top systems relying on large proprietary models with retrieval augmentation~\citep{corbeil2024iryonlp} and prompt optimization~\citep{toma2024wanglab}. How and where LLMs fail at this task remains largely unexamined.

LLMs are appealing for clinical deployment because they can be hosted locally, avoiding the need to transmit patient data to external vendors. Standard evaluation suggests that such models perform reasonably well on clinical error detection. Yet aggregate metrics evaluate each note in isolation and can be inflated by bidirectional prediction bias: a model that always predicts error (yes-bias) or always predicts no error (no-bias) can achieve high aggregate scores without distinguishing erroneous notes from correct ones.

\begin{figure}[t]
\centering
\includegraphics[width=0.88\textwidth]{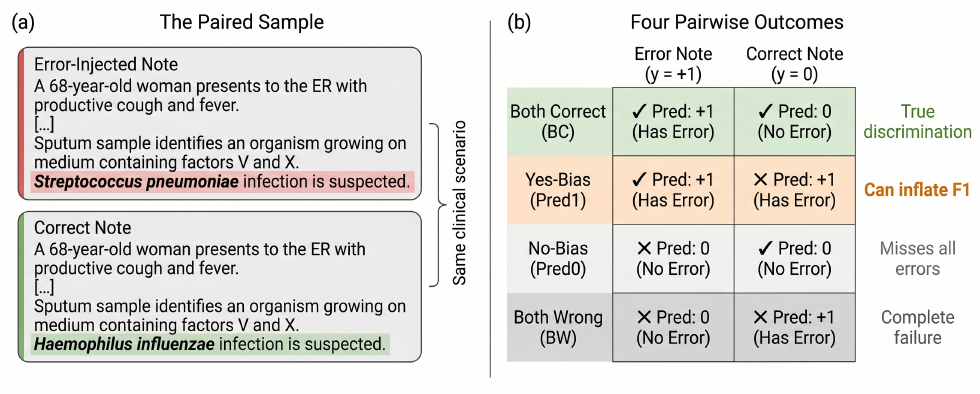}
\caption{Paired clinical error detection. (a) Each clinical scenario produces an error-injected note and a correct note differing by one sentence. (b) A model's predictions on the pair fall into four mutually exclusive outcomes.}
\label{fig:paired_sample}
\end{figure}

The paired structure of clinical error detection datasets makes this problem visible. Clinical scenarios included in these benchmarks produce two notes that differ by a single sentence: one with an injected error and one without (Figure~\ref{fig:paired_sample}). A model that truly discriminates should correctly identify the error in the first note and clear the second; a biased model assigns the same label to both, revealing that its aggregate performance reflects a default tendency rather than clinical understanding. The separation of true discrimination from response bias is long established in signal detection theory~\citep{green1966sdt}, and evaluation on minimally contrastive samples is well established through contrast sets~\citep{gardner2020evaluating}. To the best of our knowledge, clinical error detection has not adopted pairwise evaluation.

Based on this principle, we define the Both-Correct Rate (BCR), which adapts contrast consistency to clinical error detection by requiring correct classification of both members of each pair. To diagnose \emph{where} discrimination breaks down, we introduce Evidence Contrastive Analysis (ECA), which checks whether the evidence a model cites on a failed pair overlaps with the ground-truth error and correction sentences. The framework applies to any (correct, corrupted) pair, including substitution, insertion, and omission; our experiments focus on substitution-form errors because these are the errors released by the public paired benchmarks, a data limitation discussed in Section~\ref{sec:limitations}. We apply this framework to 15 LLMs across four paired clinical error-detection datasets in three languages, with four prompt configurations per model. We assess model outputs through three comprehensive strategies: aggregate pointwise metrics, pairwise \bcr{}, and \eca{} on the dominant failure mode.

In this paper, we make three contributions:

\begin{itemize}
\item \textbf{Discrimination failure is pervasive despite adequate aggregate performance.} We show that the vast majority of tested models fall below 25\% random pairwise discrimination despite achieving reasonable aggregate scores. The underlying prediction bias is bidirectional and language-dependent: the same model can exhibit yes-bias on one language and no-bias on another.

\item \textbf{Models locate error-relevant evidence but cannot judge it.} Through ECA, we show that models locate error-relevant text yet produce the same verdict on both members of the pair. In the majority of yes-bias failures, the model locates the error sentence on the erroneous note, yet labels both as erroneous.

\item \textbf{Standard metrics are structurally misleading on paired clinical data.} We show that F1 and BCR can be driven in opposite directions by the same underlying prediction bias, so that the models ranked highest by F1 are typically ranked lowest by BCR.
\end{itemize}

\section{Related Work}
\label{sec:related}

\subsection{Error Detection in Clinical Texts}
\label{sec:related:erdet}
 
The need for safer clinical documentation has motivated a series of benchmarks for automated error detection. The MEDIQA-CORR 2024 shared task formalized clinical error detection as a community challenge \citep{BenAbacha2024_MEDIQA-CORR}, with \medec{} providing a larger public corpus \citep{BenAbacha2025_MEDEC}. Subsequent work expanded the scope to multilingual evaluation \citep{Iwase2025_MedRECT, ma2026mederrbench}. Despite growing benchmark diversity, the evaluation paradigm remains focused on comparing aggregate accuracy across models and strategies. Standard benchmarks can create an evaluation illusion that obscures real-world clinical limitations \citep{Agrawal2025_evaluation}, and retrieval proficiency does not necessarily predict operational success \citep{kanithi2026medic}. Whether high scores on clinical error detection benchmarks reflect discriminative ability rather than systematic prediction bias remains unexamined. This paper addresses that gap by applying pairwise evaluation to clinical error detection, adapting the contrast consistency principle \citep{gardner2020evaluating} to the paired structure already provided by \medec{}-style datasets.

\subsection{Prediction Bias in Healthcare}
\label{sec:related:bias}

Systematic prediction bias, defined as the tendency to default to a single output class regardless of the input, is well-documented in LLMs. Sycophancy, in which models align with perceived user expectations \citep{sharma2024sycophancy}, represents one mechanism that can produce such bias; in clinical error detection, it manifests as yes-bias, defaulting to ``error present.'' In the medical domain specifically, LLMs exhibit heightened susceptibility to cognitive biases, with less capable models showing larger accuracy degradation under biased prompting \citep{schmidgall2024bias}. In parallel, a growing body of work has examined demographic bias in medical LLMs, showing that outputs in clinical decision support shift with patients' protected attributes and that prompt phrasing modulates the observed patterns \citep{poulain2026bias}; such evaluations have been scaled through automatically generated, evidence-grounded test cases \citep{fayyaz2024scalable}, and \citet{adiba2025scoping} surveys the area. This demographic line of work concerns which patients a model treats differently, whereas prediction bias concerns which output class a model defaults to. Despite these findings, prior work has predominantly characterized prediction bias as unidirectional, assuming models consistently favor one class. Whether the bias direction is stable across languages and prompt configurations has not been investigated.

\subsection{Pairwise Evaluation and Representation Gaps}
\label{sec:related:pairwise}

Signal detection theory separates a system's ability to discriminate
between classes from its response bias toward one class \citep{green1966sdt}. Because measured accuracy reflects both, response
bias alone can produce high accuracy when the class prior is skewed or
evaluation is restricted to a single class. A parallel concern motivates minimal-pair evaluation \citep{warstadt2020blimp} and contrast sets \citep{gardner2020evaluating}, which evaluate models on minimally contrastive inputs to reveal whether they have learned the relevant distinction or merely exploit superficial correlations. Despite the paired structure built into these datasets \citep{BenAbacha2024_MEDIQA-CORR, Iwase2025_MedRECT, ma2026mederrbench}, clinical NLP benchmarks have not adopted contrastive evaluation.
 
A separate line of work has shown that LLMs can encode correct information internally while producing incorrect outputs. Probing studies demonstrate that truthfulness is represented in hidden states even when generated text is wrong \citep{burns2023discovering, orgad2025llms}. Separately, chain-of-thought explanations can misstate the factors actually driving a prediction \citep{turpin2023unfaithful}. Probing requires weight access, which is unavailable for the closed models that dominate clinical deployment. Whether correct evidence can surface in a model's generated text alongside an incorrect verdict has not been systematically studied.

\section{Methods}
\label{sec:methods}

\begin{figure}[H]
\centering
\includegraphics[width=0.88\textwidth]{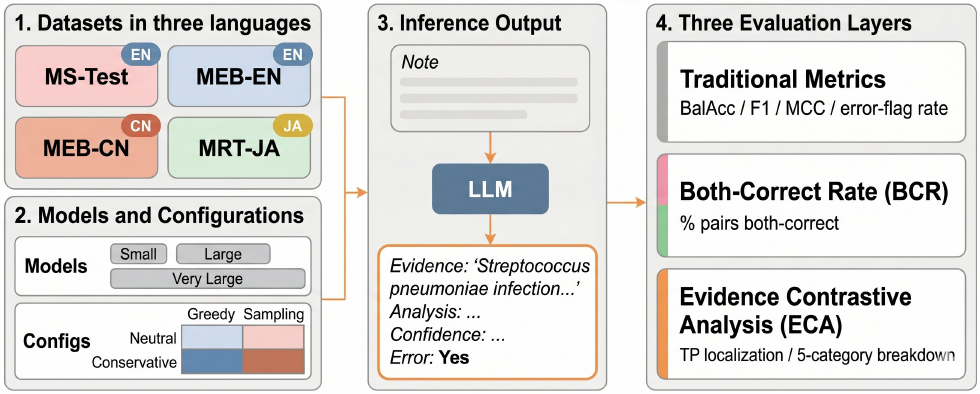}
\caption{Our framework for evaluating LLMs in clinical error detection tasks. Four paired clinical error-detection test sets across three languages, LLMs across three size tiers, and four prompt-decoding configurations are used. Each model's structured per-note output is then assessed through three parallel evaluation layers: traditional pointwise metrics, pairwise Both-Correct Rate (BCR), and Evidence Contrastive Analysis (ECA).}
\label{fig:pipeline}
\end{figure}

Pairing is the organizing principle of our evaluation (Figure~\ref{fig:paired_sample}): each error-injected note has a correct counterpart from the same clinical scenario, making the pair, not the single note, the unit at which true discrimination can be separated from response bias. Building on this, we develop three parallel evaluation layers (Figure~\ref{fig:pipeline}): traditional pointwise metrics (\S\ref{sec:eval:trad}), the Both-Correct Rate at the pair level (\S\ref{sec:eval:bcr}), and Evidence Contrastive Analysis within failed pairs (\S\ref{sec:eval:eca}). We apply this framework to 15 instruction-tuned LLMs across four test sets in three languages, under four prompt--decoding configurations per model--dataset combination.

\subsection{Datasets and Task}
\label{sec:datasets}
 
We apply our framework to four clinical error detection test sets spanning three languages (Table~\ref{tab:datasets}). MEDEC MS-Test (MS-Test) is drawn from the MEDEC MS corpus \citep{BenAbacha2025_MEDEC}; MedErrBench-EN (MEB-EN) and MedErrBench-CN (MEB-CN) are the English and Chinese test sets of MedErrBench~\citep{ma2026mederrbench}; and MedRECT-JA (MRT-JA) is the Japanese test set of MedRECT, with nine error types~\citep{Iwase2025_MedRECT}.
 
In each dataset, error-injected notes are paired with correct counterparts from the same clinical scenarios. Whether these pairs are released explicitly varies by dataset, and not every released note participates in a matched pair; we apply dataset-specific procedures to identify usable pairs, with details in Appendix~\ref{app:pairs}.

MedRECT-JA warrants a caveat: its 190 error notes share only 105 unique clean notes, a many-to-one structure that may inflate within-pair error correlation (Appendix~\ref{app:pairs}).
 
We restrict evaluation to the binary error-flag sub-task: given a single clinical note, decide whether it contains a medical error, thereby measuring pairwise discrimination without conflating detection with correction. Notes are evaluated independently; the model never sees a pair side-by-side, so pairwise discrimination must be inferred from each note's content alone, with pairing applied post-hoc to compute BCR.

\begin{table}[t]
\caption{Datasets used for inference and evaluation. Error and Clean are the sub-counts of notes with and without errors, respectively; Pairs is the number of matched (error, correct) tuples used for BCR evaluation, which can differ from $\min(\text{Error}, \text{Clean})$ when notes lack a counterpart or share counterparts across pairs.}
\label{tab:datasets}
\centering
\small
\begin{tabular}{lcccccc}
\toprule
Dataset & Samples & Error & Clean & Pairs & Language & Source \\
\midrule
MS-Test & 597 & 311 & 286 & 286 & English &
  \citet{BenAbacha2025_MEDEC} \\
MEB-EN & 208 & 104 & 104 & 104 & English &
  \citet{ma2026mederrbench} \\
MEB-CN & 200 & 100 & 100 & 100 & Chinese &
  \citet{ma2026mederrbench} \\
MRT-JA & 295 & 190 & 105 & 190 & Japanese &
  \citet{Iwase2025_MedRECT} \\
\bottomrule
\end{tabular}
\end{table}

\subsection{Models}
\label{sec:models}
 
We evaluate 15 instruction-tuned LLMs (Appendix~\ref{app:models}, Table~\ref{tab:models:apx}) across three size tiers (3--8B, 27--32B, and 70B) and five model families; five are medical-domain, the other ten general-purpose. This grid enables two comparisons: scale within a family (e.g., Qwen 3 at 4B, 8B, and 32B) and medical specialization at matched scale (e.g., Gemma 3-27B vs.\ MedGemma 27B).
 
Clinical sites adopting LLMs for error detection are unlikely to curate task-specific few-shot demonstrations, and a fair comparison across model families with different training paradigms requires a common starting point. We therefore evaluate all models zero-shot, without task-specific examples or fine-tuning.

\paragraph{Precision.} Models at 27B parameters and below are loaded in native bfloat16 (bf16); the two 70B models are loaded in 8-bit floating point (fp8) via vLLM~\citep{kwon2023vllm} due to memory constraints, which may slightly reduce their performance relative to full precision.

\subsection{Prompt Design and Decoding}
\label{sec:prompts}
 
We treat prompt and decoding choices as a 2$\times$2 perturbation matrix that probes two orthogonal sources of variability: whether instruction wording can shift prediction bias, and whether sampling noise drives the observed patterns. Reporting the cross-configuration mean and standard deviation across the four resulting configurations lets us separate stable model behavior from configuration-driven artifacts, a robustness check that single-configuration benchmarks cannot provide.
 
The \emph{neutral} prompt instructs the model to act as a skilled medical doctor performing a standard clinical review. The \emph{conservative} prompt adds an explicit instruction to prefer ``no error'' when the evidence is ambiguous, testing whether directed caution can shift prediction bias without improving discrimination. Both prompts require a structured four-line output (\texttt{Evidence}, \texttt{Analysis}, \texttt{Confidence}, \texttt{Error:Yes/No}), providing both the binary verdict for evaluation and the textual evidence used by \eca{} (Section~\ref{sec:eval:eca}). Dataset-specific error type lists are provided in the native language of each dataset.
 
For decoding, we use greedy decoding as the primary setting because it yields fully reproducible outputs and isolates bias effects from sampling noise, and complement it with stochastic sampling to test whether sampling variability shifts the observed bias patterns. Full prompt templates and exact decoding parameters appear in Appendix~\ref{app:prompts}.

\subsection{Evaluation Framework}
\label{sec:eval}
 
\subsubsection{Traditional Metrics}
\label{sec:eval:trad}
 
We report balanced accuracy, F1, precision, recall, specificity, and the Matthews correlation coefficient (MCC). We additionally report the \emph{error-flag rate}: the fraction of notes for which the model outputs \texttt{Error:Yes}. Balanced accuracy and MCC serve as primary anchors because they account for class imbalance and penalize degenerate classifiers, respectively. F1 and recall are reported to expose the discrepancy between bias-sensitive and bias-resistant metrics.
 
\subsubsection{Both-Correct Rate (BCR)}
\label{sec:eval:bcr}
 
BCR adapts the contrast consistency principle
\citep{gardner2020evaluating} to clinical error detection. For each pair $(x_e, x_c)$ consisting of an error-injected and a correct note from the same clinical scenario, the model's predictions $(\hat{y}_e, \hat{y}_c)$ fall into four mutually exclusive categories. We use the shorthand \textit{Pred1} (both predicted erroneous) to denote pairs where the model flags both members as containing an error, and \textit{Pred0} (both predicted non-erroneous) for pairs where the model judges both error-free. These two outcomes correspond to systematic yes-bias and no-bias failure modes, respectively.

\begin{enumerate}
\item \textbf{Both Correct (BC):} $\hat{y}_e = 1$ and $\hat{y}_c = 0$. Correct on both members of the pair.
\item \textbf{\textit{Pred1}:} $\hat{y}_e = 1$ and $\hat{y}_c = 1$. Both members predicted as containing an error (yes-bias).
\item \textbf{\textit{Pred0}:} $\hat{y}_e = 0$ and $\hat{y}_c = 0$. Both members predicted as error-free (no-bias).
\item \textbf{Both Wrong (BW):} $\hat{y}_e = 0$ and $\hat{y}_c = 1$. Inverted judgment on both members.
\end{enumerate}
 
BCR is the fraction of pairs classified as Both Correct:
\begin{equation}
\mathrm{BCR} = \frac{1}{N}\sum_{i=1}^{N}\mathbf{1}[\hat{y}_{e,i}=1
\;\wedge\; \hat{y}_{c,i}=0].
\label{eq:bcr}
\end{equation}

The population counterpart of Equation~\ref{eq:bcr} is
$\mathbb{E}[\mathrm{BCR}] = P(\hat{y}_e = 1, \hat{y}_c = 0 \mid y_e = 1, y_c = 0)$: the joint probability that a random pair is classified correctly on both members. Under the null hypothesis that within-pair predictions are conditionally independent given the pair labels, with each prediction depending only on its own note, this joint factorizes into marginals, giving
\begin{equation}
\mathbb{E}[\mathrm{BCR}]_{\mathrm{indep}}
= P(\hat{y}_e = 1 \mid y_e = 1) \cdot P(\hat{y}_c = 0 \mid y_c = 0)
= \mathrm{sensitivity} \cdot \mathrm{specificity}.
\label{eq:bcr-indep}
\end{equation}
We define the \emph{independence ratio} as
\begin{equation}
R_{\text{independence}} = \frac{\mathrm{BCR}}{\mathbb{E}[\mathrm{BCR}]_{\mathrm{indep}}},
\label{eq:bcr-ratio}
\end{equation}
which measures how far observed BCR departs from the independence baseline implied by the model's own marginal sensitivity and specificity. A consistent $R_{\text{independence}} < 1$ indicates that within-pair prediction outcomes are systematically dependent: joint success on both members of a pair occurs less often than within-pair independence would predict. We emphasize that our contribution is this diagnostic apparatus---the independence ratio, which isolates within-pair dependence that contrastive accuracy alone does not expose, together with \eca{} (Section~\ref{sec:eval:eca}), which localizes \emph{where} discrimination breaks down---rather than the \bcr{} statistic itself, which is a direct adaptation of contrast consistency~\citep{gardner2020evaluating}.

Beyond this stochastic baseline, \bcr{} obeys a deterministic upper bound. Because a Both-Correct pair requires $\hat{y}_e = 1$ \emph{and} $\hat{y}_c = 0$, the Both-Correct event is contained in each single-note success event, so
\begin{equation}
\mathrm{BCR} \leq \min(\mathrm{sensitivity}, \mathrm{specificity}).
\label{eq:bcr-bound}
\end{equation}
Any class bias therefore caps \bcr{} at the weaker marginal, however high F1 climbs through recall on the favored class, so the F1--\bcr{} divergence in Section~\ref{sec:results:deception} is structural rather than model-specific. A short proof and an always-error corollary (F1~$=2/3$ with $\mathrm{BCR}=0$ on balanced data) are given in Appendix~\ref{app:bound}.

For a random classifier with $\mathrm{sensitivity} = \mathrm{specificity} = 0.5$, Equation~\ref{eq:bcr-indep} yields a 25\% BCR baseline; this figure is the balanced special case and applies only to the three class-balanced pair sets (MS-Test, MEB-EN, MEB-CN). On MedRECT-JA (190 of 295 notes contain errors), a random predictor calibrated to the empirical class prevalence (outputting $\hat{y}=1$ with probability $190/295$ independent of the input) has $\mathrm{sensitivity} = 190/295$ and $\mathrm{specificity} = 105/295$, giving an independence baseline of $(190/295)(105/295) \approx 22.9\%$. The general, balance-free reference is therefore the independence ratio $R_{\text{independence}}$ (Equation~\ref{eq:bcr-ratio}) rather than any fixed percentage. We report the four-configuration mean $\pm$ standard deviation as the primary BCR measure; cross-configuration variation is informative in itself about prompt sensitivity.

\subsubsection{Evidence Contrastive Analysis (ECA)}
\label{sec:eval:eca}

\begin{figure}[t]
\centering 
\includegraphics[width=0.88\textwidth]{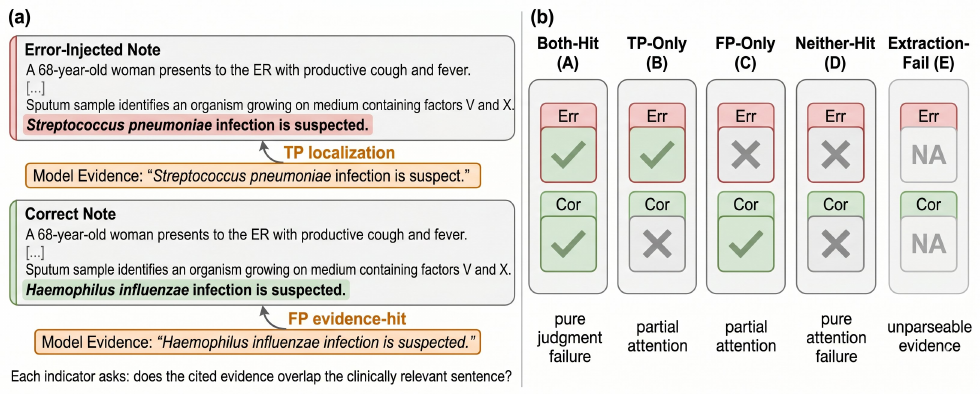}
\caption{Evidence Contrastive Analysis (\eca{}). \textbf{(a)}~The two indicators, illustrated on a shared clinical scenario: \emph{TP localization} asks whether the cited evidence on the error note (Streptococcus pneumoniae) overlaps the error sentence; \emph{FP evidence-hit} asks the same question on the correct note (Haemophilus influenzae). 
\textbf{(b)}~The five outcomes arising from combining the two indicators across a pair. Both-Hit (A) isolates a judgment failure: the model attends to the correct location in both notes, yet still cannot determine which version is incorrect. Neither-Hit (D) reflects full attention failure. TP-Only (B) and FP-Only (C) are partial-attention failures. Extraction-Fail (E) captures pairs with unparseable \texttt{Evidence} fields.}
\label{fig:eca}
\end{figure}

BCR reveals whether models discriminate between paired samples, but not which stage of their reasoning fails: locating the relevant evidence or judging it correctly once located. Evidence Contrastive Analysis (\eca{}) is a post-hoc diagnostic procedure that, for each pair, scores whether the cited \texttt{Evidence} field overlaps the clinically relevant sentence on the error note (\emph{TP localization}) and on the correct note (\emph{FP evidence-hit}). Overlap is scored by sub-string containment or $\geq 60\%$ word coverage (threshold sensitivity in Appendix~\ref{app:robustness}), adapting the rationale token-overlap convention~\citep{deyoung2020eraser}, with character-level counting for Chinese and Japanese where word boundaries are ill-defined. Combining these two indicators across the pair yields the five mutually exclusive categories shown in Figure~\ref{fig:eca}, which separate attention failures from judgment failures.

\subsubsection{Parse Failure Handling}
\label{sec:eval:parse}

Although our prompts specify a strict four-line output format, not all models achieve perfect compliance: some outputs add extra text, omit the \texttt{Error} field, or produce a malformed structure, leaving no parseable verdict. Pairs where either member cannot be parsed are excluded from BCR computation. To verify that this exclusion does not bias the results, we compare against a random-assignment imputation that fills in unparseable verdicts using the model's overall accuracy (Appendix~\ref{app:robustness}). Parse failures are themselves captured as Extraction-Fail (E) in \eca{} (Section~\ref{sec:eval:eca}).

\section{Results}
\label{sec:results}

\subsection{Traditional Metrics Suggest Moderate Performance}
\label{sec:results:trad}

Across all 240 runs (15 models $\times$ 4 datasets $\times$ 4 configurations), balanced accuracy ranges from 0.45 to 0.71, with MCC near zero for most models, indicating performance close to chance at distinguishing error-containing notes from correct ones. F1 scores, however, range up to 0.80: in 178 of 240 runs (74\%), F1 exceeds 0.5 while balanced accuracy stays at or below 0.6. Under standard evaluation practice, these F1 scores would suggest that most models achieve moderate performance on clinical error detection. Full per-run metrics for all 240 runs appear in Appendix~\ref{app:trad240} (Tables~\ref{tab:trad240:cg}--\ref{tab:trad240:ns}). The pairwise analysis that follows reveals that this apparent performance does not reflect discriminative ability at the pair level.

\subsection{Pairwise BCR Reveals Low Discrimination}
\label{sec:results:bcr}

Table~\ref{tab:bcr} presents the four-configuration mean \bcr{} for each model--dataset combination. Thirteen of 15 models fall below 25\% mean \bcr{} across datasets, the level expected from a random classifier on balanced data. Only Qwen~3-32B (28.0\%) and UltraMedical~70B (25.7\%) exceed this threshold, and both still fail to correctly classify both members of a pair more than 70\% of the time.

The 25\% baseline is the balanced special case (for prevalence-imbalanced MRT-JA it is $22.9\%$; the balance-free reference is the independence ratio of Equation~\ref{eq:bcr-ratio}, analyzed in Section~\ref{sec:results:deception}). The \emph{Sens} and \emph{Spec} columns of Table~\ref{tab:bcr} make the mechanism concrete: top-ranked models pair moderate sensitivity with moderate specificity, whereas the lowest-\bcr{} models drive one marginal to an extreme (e.g., Gemma~3-4B at 94.2\% sensitivity but 7.2\% specificity), exactly the configuration a single aggregate score conceals.

\begin{table}[h]
\caption{Four-configuration mean \bcr{} (\%) $\pm$ standard deviation by model and dataset. \emph{Sens} and \emph{Spec} are the model's mean pointwise sensitivity and specificity, averaged over all 16 runs; the highest-\bcr{} models pair moderate sensitivity with moderate specificity, while the lowest-\bcr{} models collapse one marginal toward its extreme. $\dagger$~Medical-domain model. Sorted by cross-dataset mean \bcr{}.}
\label{tab:bcr}
\centering\small\setlength{\tabcolsep}{3pt}
\begin{tabular}{lrrrrrrr}
\toprule
& \multicolumn{5}{c}{\bcr{} (\%)} & \multicolumn{2}{c}{Marginal (\%)} \\
\cmidrule(lr){2-6}\cmidrule(lr){7-8}
Model & MS-Test & MEB-EN & MEB-CN & MRT-JA & Mean & Sens & Spec \\
\midrule
Qwen 3-32B            & $22.6 {\scriptstyle\pm 3.7}$ & $30.8 {\scriptstyle\pm 11.1}$ & $32.0 {\scriptstyle\pm 5.5}$ & $26.6 {\scriptstyle\pm 1.9}$ & 28.0 & 69.4 & 51.4 \\
UltraMed-70B$\dagger$ & $24.3 {\scriptstyle\pm 9.8}$ & $43.7 {\scriptstyle\pm 2.2}$  & $16.8 {\scriptstyle\pm 10.0}$ & $18.0{\scriptstyle\pm 5.7}$ & 25.7 & 47.0 & 73.5 \\
MedGemma 27B$\dagger$ & $20.3 {\scriptstyle\pm 4.9}$ & $29.1 {\scriptstyle\pm 8.6}$  & $14.6 {\scriptstyle\pm 5.7}$ & $28.5 {\scriptstyle\pm 4.2}$ & 23.1 & 61.2 & 56.0 \\
Llama 3.1-70B         & $20.2 {\scriptstyle\pm 3.6}$ & $27.1 {\scriptstyle\pm 15.6}$ & $15.3 {\scriptstyle\pm 2.8}$ & $17.4 {\scriptstyle\pm 7.0}$ & 20.0 & 63.6 & 51.4 \\
UltraMed-8B$\dagger$  & $19.6 {\scriptstyle\pm 5.8}$ & $28.5 {\scriptstyle\pm 2.5}$  & $18.4 {\scriptstyle\pm 5.5}$ & $13.5 {\scriptstyle\pm 1.8}$ & 20.0 & 43.4 & 65.0 \\
Qwen 3-4B             & $12.0 {\scriptstyle\pm 5.9}$ & $13.2 {\scriptstyle\pm 6.6}$  & $23.0 {\scriptstyle\pm 1.2}$ & $19.9 {\scriptstyle\pm 1.9}$ & 17.0 & 83.5 & 27.8 \\
MedGemma 4B$\dagger$  & $16.0 {\scriptstyle\pm 3.8}$ & $16.6 {\scriptstyle\pm 6.6}$  & $14.8 {\scriptstyle\pm 2.7}$ & $12.5 {\scriptstyle\pm 6.6}$ & 15.0 & 66.5 & 38.6 \\
Llama 3.1-8B          & $14.1 {\scriptstyle\pm 7.5}$ & $16.9 {\scriptstyle\pm 13.1}$ & $14.6 {\scriptstyle\pm 4.6}$ & $12.8 {\scriptstyle\pm 7.4}$ & 14.6 & 70.9 & 34.5 \\
Phi-4-mini            & $13.9 {\scriptstyle\pm 7.7}$ & $12.7 {\scriptstyle\pm 5.6}$  & $ 7.5 {\scriptstyle\pm 4.8}$ & $16.5 {\scriptstyle\pm 1.5}$ & 12.7 & 86.1 & 18.4 \\
Qwen 3-8B             & $11.2 {\scriptstyle\pm 3.2}$ & $13.0 {\scriptstyle\pm 1.4}$  & $ 8.0 {\scriptstyle\pm 3.2}$ & $16.1 {\scriptstyle\pm 4.6}$ & 12.1 & 56.4 & 49.4 \\
MediPhi 4B$\dagger$      & $11.5 {\scriptstyle\pm 6.5}$ & $10.3 {\scriptstyle\pm 4.6}$  & $11.2 {\scriptstyle\pm 6.1}$ & $ 9.7 {\scriptstyle\pm 2.1}$ & 10.7 & 87.4 & 15.4 \\
Gemma 3-27B           & $10.8 {\scriptstyle\pm 5.7}$ & $ 9.9 {\scriptstyle\pm 8.4}$  & $13.2 {\scriptstyle\pm 8.3}$ & $ 7.1 {\scriptstyle\pm 3.3}$ & 10.2 & 85.2 & 21.0 \\
Llama 3.2-3B          & $ 5.9 {\scriptstyle\pm 3.1}$ & $ 8.4 {\scriptstyle\pm 4.2}$  & $16.3 {\scriptstyle\pm 7.7}$ & $10.0 {\scriptstyle\pm 7.1}$ & 10.2 & 84.1 & 17.5 \\
Mistral 7B            & $ 4.5 {\scriptstyle\pm 3.4}$ & $ 5.0 {\scriptstyle\pm 3.6}$  & $ 6.3 {\scriptstyle\pm 5.4}$ & $12.8 {\scriptstyle\pm 6.4}$ &  7.1 & 88.8 & 10.9 \\
Gemma 3-4B            & $ 5.5 {\scriptstyle\pm 3.8}$ & $ 2.2 {\scriptstyle\pm 2.4}$  & $ 4.2 {\scriptstyle\pm 3.3}$ & $ 6.3 {\scriptstyle\pm 1.1}$ &  4.6 & 94.2 &  7.2 \\
\bottomrule
\end{tabular}
\end{table}

Two secondary trends emerge within this pattern. First, scaling within model families yields modest gains: Qwen improves from 17.0\% (4B) to 28.0\% (32B), though non-monotonically (12.1\% at 8B), and Llama from 10.2\% (3B) to 20.0\% (70B). However, these gains are concentrated on English datasets; for example, Llama 3B-to-70B gains 18.7~pp on MedErrBench-EN but decreases by 1.0~pp on MedErrBench-CN. Second, medical-domain pretraining often provides larger gains than scaling alone: Gemma~3-27B (general-purpose) achieves 10.2\% mean \bcr{}, while MedGemma~27B (medical) reaches 23.1\%, more than doubling its general-purpose counterpart. This holds for four of five matched general–medical pairs; MediPhi 4B (10.7\%) is the exception, underperforming Phi-4-mini (12.7\%).

Cross-configuration standard deviations range from 1.1~pp to 15.6~pp across individual model--dataset entries. Llama~3.1-70B is the most unstable overall. On MedErrBench-EN, its \bcr{} ranges from 9.0\% to 45.2\% across the four configurations, a spread of 36.2~pp.

\subsection{Bias Is Bidirectional and Language-Dependent}
\label{sec:results:bias}

The bias patterns underlying these \bcr{} results are not uniform across languages. Table~\ref{tab:bias} reports the mean error-flag rate for each model--dataset combination, along with the resulting bias category. Eight of 15 models change bias category across datasets, and for seven of these, the mean error-flag rate falls on opposite sides of parity on different datasets; Qwen~3-8B reverses outright, from no-bias on Chinese (0.36) to yes-bias on Japanese (0.69). UltraMedical~8B illustrates the more common pattern: balanced prediction rates on English datasets (0.42--0.55) shifting to strong no-bias on Chinese and Japanese (0.27--0.32).

Several models also switch bias direction outright between prompt variants: on at least one dataset, seven of 15 models show yes-bias under neutral prompting but no-bias under conservative prompting (per-configuration error-flag rates in Appendix~\ref{app:detailed}, Tables~\ref{tab:trad240:cg}--\ref{tab:trad240:ns}), so conservative prompting does not simply shift prediction rates toward balance.

\definecolor{yesbg}{RGB}{253,224,224}            
\definecolor{nobg}{RGB}{224,234,253} 
\begin{table}[t] 
\caption{Mean error-flag rate per model--dataset combination.\colorbox{yesbg}{Pink}~cells indicate yes-bias (rate~$\geq 0.6$); \colorbox{nobg}{blue}~cells indicate no-bias (rate~$<0.4$); unshaded cells are balanced ($0.4 \leq$~rate~$<0.6$). A check in the Shift column marks models whose bias category changes across datasets. $\dagger$~Medical-domain model.} 
\label{tab:bias}
\centering\small\setlength{\tabcolsep}{4pt}
\begin{tabular}{lccccc} 
\toprule Model & MS-Test & MEB-EN & MEB-CN & MRT-JA & Shift? \\ 
\midrule 
Llama 3.2-3B          & \cellcolor{yesbg}.94 & \cellcolor{yesbg}.88 & \cellcolor{yesbg}.63 & \cellcolor{yesbg}.89 &            \\ 
Gemma 3-4B            & \cellcolor{yesbg}.94 & \cellcolor{yesbg}.98 & \cellcolor{yesbg}.89 & \cellcolor{yesbg}.93 &            \\ 
Qwen 3-4B             & \cellcolor{yesbg}.81 & \cellcolor{yesbg}.91 & \cellcolor{yesbg}.61 & \cellcolor{yesbg}.80 &            \\ 
MedGemma 4B$\dagger$  & \cellcolor{yesbg}.60 & \cellcolor{yesbg}.66 & .50 & \cellcolor{yesbg}.80 & \checkmark \\ 
MediPhi 4B$\dagger$   & \cellcolor{yesbg}.84 & \cellcolor{yesbg}.89 & \cellcolor{yesbg}.85 & \cellcolor{yesbg}.86 &            \\ 
Phi-4-mini            & \cellcolor{yesbg}.79 & \cellcolor{yesbg}.89 & \cellcolor{yesbg}.88 & \cellcolor{yesbg}.81 &            \\ 
Mistral 7B            & \cellcolor{yesbg}.95 & \cellcolor{yesbg}.96 & \cellcolor{yesbg}.93 & \cellcolor{yesbg}.72 &            \\ 
Llama 3.1-8B          & \cellcolor{yesbg}.67 & \cellcolor{yesbg}.78 & .43 & \cellcolor{yesbg}.85 & \checkmark \\ 
UltraMed-8B$\dagger$  &                 .42 &                  .55 & \cellcolor{nobg}.32  & \cellcolor{nobg}.27  & \checkmark \\ 
Qwen 3-8B             &                 .50 & \cellcolor{yesbg}.60 & \cellcolor{nobg}.36  & \cellcolor{yesbg}.69 & \checkmark \\ 
\midrule 
Gemma 3-27B           & \cellcolor{yesbg}.67 & \cellcolor{yesbg}.92 & \cellcolor{yesbg}.77 & \cellcolor{yesbg}.92 &            \\
MedGemma 27B$\dagger$ &                 .45 & \cellcolor{yesbg}.60 & .43 & \cellcolor{yesbg}.65 & \checkmark \\
Qwen 3-32B            & \cellcolor{yesbg}.62 & \cellcolor{yesbg}.78 & .43 &                  .56 & \checkmark \\
\midrule 
Llama 3.1-70B         &                 .50 & \cellcolor{yesbg}.78 & .41 &                  .57 & \checkmark \\
UltraMed-70B$\dagger$ & \cellcolor{nobg}.36 &                  .59 & \cellcolor{nobg}.20  & \cellcolor{nobg}.33  & \checkmark \\
\bottomrule
\end{tabular}
\end{table}

\subsection{The Localization-Judgment Gap}
\label{sec:results:eca}

Having established that discrimination fails and that yes-bias (\textit{Pred1}) is the dominant failure mode across our experiments (for the two no-bias models, UltraMedical~8B and 70B, \textit{Pred0} dominates instead), we next examine whether models nonetheless attend to error-relevant content in their outputs. We focus \eca{} on \textit{Pred1} pairs, which concentrate the bulk of failures and therefore offer the greatest statistical power for diagnosing what happens when models fail pairwise discrimination while still producing reasoning evidence.

\begin{figure}[h] 
\centering
\includegraphics[width=0.88\textwidth]{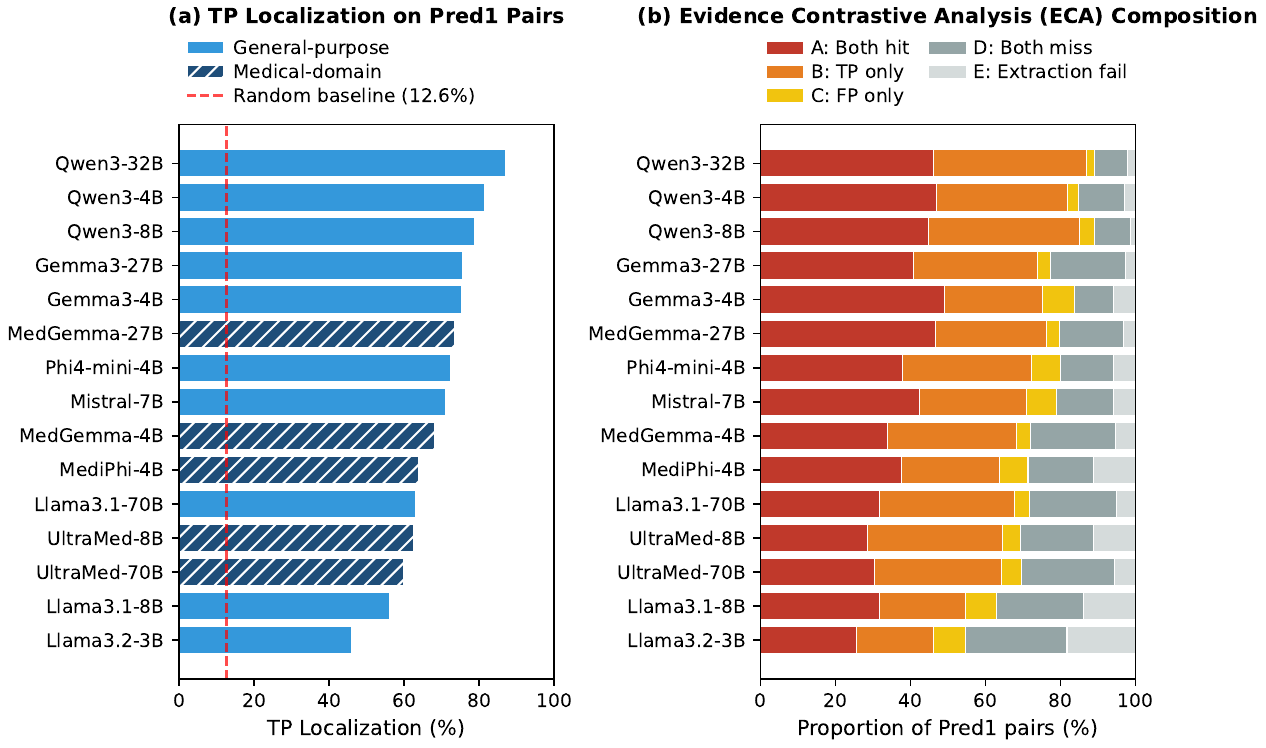}
\caption{The localization-judgment gap on MS-Test (four-configuration mean across 15 models). (a)~TP localization per model; dashed line is the random baseline (12.6\%). (b)~\eca{} category composition per model, as percentage of \textit{Pred1} pairs; category definitions in Figure~\ref{fig:eca}.} 
\label{fig:ljg}
\end{figure}

TP localization measures whether the evidence cited by the model on the error note overlaps with the actual error sentence. Random baselines, corresponding to a sentence-level uniform pick, are the inverse of the average number of sentences per note (12.6\% for MS-Test, 9.3\% for MEB-EN, 27.7\% for MEB-CN, 10.1\% for MRT-JA). On MS-Test, all 15 models exceed the random baseline under the four-configuration mean (Figure~\ref{fig:ljg}a). Averaged across the 15 models, per-dataset TP localization rates are 87\% on MEB-EN, 70\% on MEB-CN, 69\% on MS-Test, and 42\% on MRT-JA (full per-model values in Appendix~\ref{app:eca:full}, Table~\ref{tab:eca:full}).

The \eca{} category breakdown (Figure~\ref{fig:ljg}b) shows that correct localization on the error note coexists with pair failure. On MS-Test, Both-Hit (A) and TP-Only (B) together, which both indicate that the model attended to the error sentence on the error note, account for 46--87\% of categorized \textit{Pred1} pairs across models (mean 70\%); the model thus locates the right content in most failed pairs, yet still misjudges them. Within this, Both-Hit (A) alone captures the purest form of the localization-judgment gap: the model attends to the relevant sentence on \emph{both} notes yet judges both as erroneous (26--49\% of \textit{Pred1} pairs, mean 38\%). Neither-Hit (D), which represents full attention failure, accounts for only 18\% on average.

\begin{figure}[t]
\centering  
\includegraphics[width=0.88\textwidth]{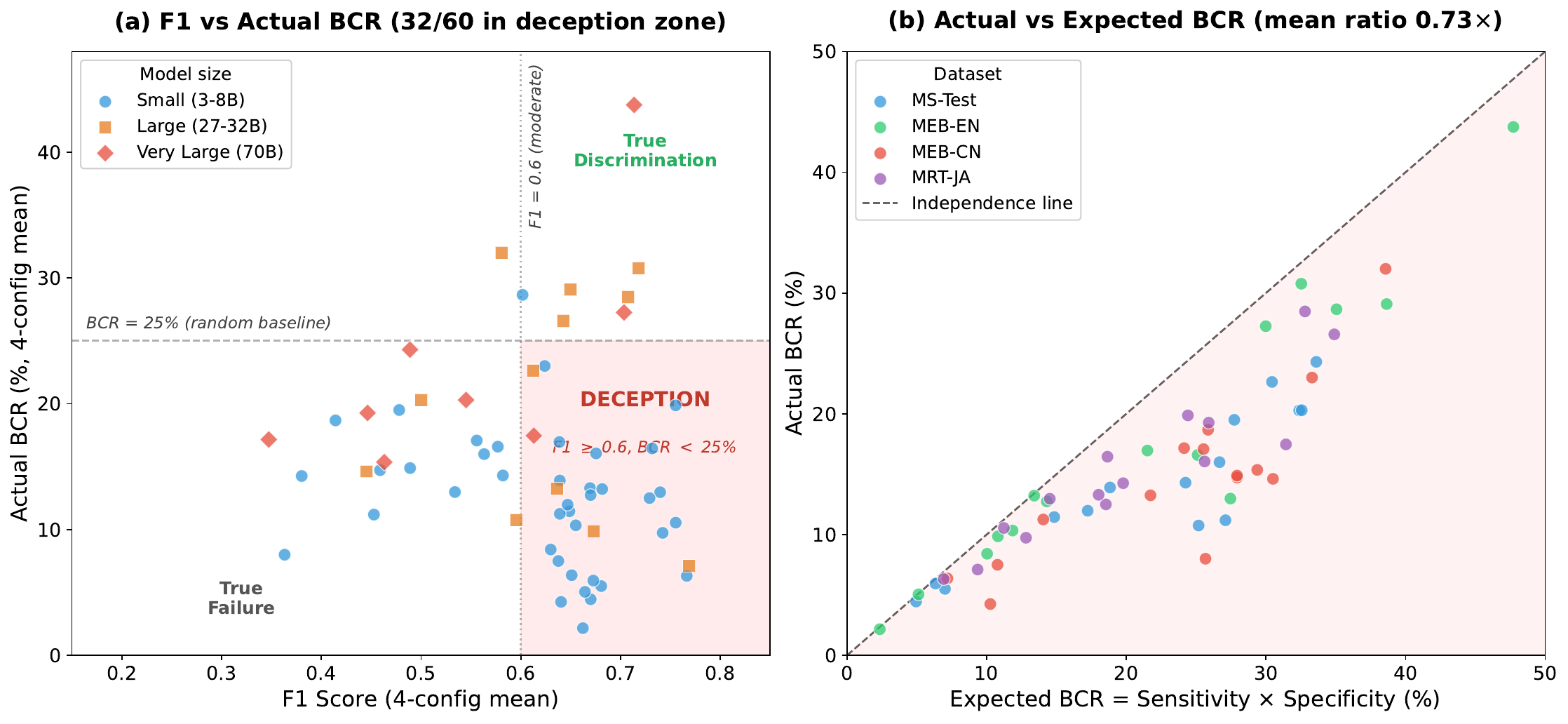}
\caption{Metric deception across 60 model--dataset entries (four-configuration means). (a)~Shaded region indicates F1~$\geq$~0.6 with \bcr{}~$<$~25\%. (b)~Actual \bcr{} vs.\ expected \bcr{} under statistical independence (Equation~\ref{eq:bcr-indep}); the dashed line is the independence baseline ($R_{\text{independence}} = 1$), below which all 60 entries fall.}
\label{fig:deception}
\end{figure}

\subsection{Prediction Bias Mediates the Relationship Between F1 and BCR}
\label{sec:results:deception}

The preceding sections raise a question: if models fail at pairwise discrimination, why do traditional metrics suggest otherwise? Figure~\ref{fig:deception} addresses this by examining how prediction bias relates to both F1 and \bcr{}. We operationalize prediction bias as the error-flag rate (Section~\ref{sec:eval:trad}).

The error-flag rate simultaneously inflates F1 ($r = +0.85$ across 60 model--dataset entries) and suppresses \bcr{} ($r = -0.49$), while the direct F1--\bcr{} correlation is near zero ($r = -0.06$). The two mechanisms pull F1 and \bcr{} in opposite directions, so the direct F1--\bcr{} association is dataset-dependent rather than uniform: it is negative on MS-Test ($r = -0.66$), MEB-CN ($r = -0.14$) and MRT-JA ($r = -0.23$), positive on MEB-EN ($r = +0.31$), and averages to near zero once the four datasets are pooled (Appendix~\ref{app:mediation}, Table~\ref{tab:mediation}). This opposition is the empirical face of the structural bound of Equation~\ref{eq:bcr-bound}: a collapsed marginal caps \bcr{} while F1 is free to rise. The practical consequence is visible in the rankings rather than in the pooled correlation: on three of four datasets, the top-3 models by F1 and the top-3 by \bcr{} share zero overlap (Appendix~\ref{app:mediation}, Table~\ref{tab:top3}).

Thirty-two of 60 entries (53\%) fall in a deception zone where F1~$\geq$~0.6 but \bcr{}~$<$~25\% (Figure~\ref{fig:deception}a). On MS-Test, the three models with the highest F1 (Gemma~3-4B, Llama~3.2-3B, Mistral~7B, all with F1~$>$~0.65) have the three lowest \bcr{} values (4.5--5.9\%).

The independence ratio~(Equation~\ref{eq:bcr-ratio}) averages $0.73$ across the 60 model--dataset entries, with all 60 falling below the independence line (Figure~\ref{fig:deception}b), confirming that within-pair errors are systematically correlated. Per-dataset ratios range from $0.62$ on MEB-CN (strongest correlation) to $0.84$ on MEB-EN (closest to independence); full details in Appendix~\ref{app:mediation}.

\paragraph{A proprietary reference point.} Our systematic evaluation covers open-weight models, which are the systems that most healthcare organizations can deploy on-premise for PHI reasons. To situate that range against a frontier proprietary system, we additionally ran GPT-5~mini\footnote{OpenAI API model \texttt{gpt-5-mini-2025-08-07}.} on MS-Test within our available budget. As a reasoning model, it does not accept a temperature setting, so we vary only the prompt (neutral and conservative), running two runs. Its \bcr{} of 42.3\% (F1~$=0.66$, balanced accuracy~$=0.69$) exceeds every open-weight model on MS-Test (best 24.3\%, UltraMedical~70B) and sits outside the deception zone; its observed \bcr{} still falls below its own independence baseline of 47.1\% ($\mathrm{sensitivity}\times\mathrm{specificity}$), consistent with the within-pair dependence seen throughout. Training data is undisclosed for GPT-5~mini and open-weight models alike, so benchmark exposure cannot be ruled out for either. Because contamination inflates rather than depresses scores, it bears on a strong result like this one, not on our main finding of pervasive failure of discrimination. We read this as one illustrative data point, not a full proprietary evaluation.

\section{Discussion}
\label{sec:discussion}

\subsection{Implications for Clinical Model Selection}
\label{sec:disc:selection}
 
Thirteen of 15 models fall below 25\% random \bcr{} (Section~\ref{sec:results:bcr}), yet many of these same models achieve F1 scores above 0.6. F1-based model selection would therefore systematically favor the weakest discriminators. In a clinical setting, evaluating candidate models by F1 on MS-Test, Gemma~3-4B, Llama~3.2-3B, Mistral~7B would rank as the top three, precisely the models with the lowest \bcr{} (4.5--5.9\%). This is not a hypothetical concern but the default outcome of current evaluation practice.
 
This vulnerability is specific to the accuracy regime in which clinical error detection operates. At high balanced accuracy (e.g., $\approx$90\%), bias has limited room to inflate F1, and \bcr{} is mathematically constrained to be high. Clinical error detection, however, operates at $\approx$50--60\% balanced accuracy, where a model can achieve F1~$\approx$~0.68 by always predicting a single class. In this regime, pairwise evaluation is not a refinement but the only way to separate discriminative ability from systematic bias. Many clinical NLP tasks operate at similar accuracy levels and face the same risk. This concern is consistent with a growing literature showing that aggregate confusion-matrix metrics can reward degenerate prediction strategies, remain high under class imbalance or weak discrimination, and are formally improper as decision measures~\citep{lipton2014thresholding, reinke2024pitfalls, vancalster2025landig}.
 
That few model--dataset entries achieve both high F1 and high \bcr{} is an empirical finding, not a mathematical necessity. If models achieved high F1 through correct pairwise classification rather than bias, the two metrics would correlate positively. The almost empty upper-right quadrant of Figure~\ref{fig:deception}a confirms that, across most models tested, high F1 is largely achieved through bias alone. For safety-critical clinical NLP, paired metrics such as \bcr{} should be reported alongside traditional metrics.

Concretely, paired evaluation serves three points in the deployment pipeline. \emph{Benchmark designers} can report \bcr{} and the independence ratio next to F1 and MCC, so that published rankings reflect discrimination rather than a default class tendency. \emph{Healthcare-organization governance and procurement teams}, who are often limited to open-weight, on-premise models by PHI and data-governance rules, can screen candidate models on paired data as a gate before adoption, catching systems that pass on aggregate metrics but fail pairwise. \emph{Vendors} can run the same check before release. This screen is a necessary first step, not a replacement for prospective, workflow-level evaluation with clinicians in the loop, nor for broader error-type coverage (Section~\ref{sec:limitations}). What it establishes is a precondition: whether a model can tell an erroneous note from its clean counterpart at all. Most of the models we tested cannot.

\subsection{Evidence Production and Clinical Judgment}
\label{sec:disc:ljg}

The localization-judgment gap documented in Section~\ref{sec:results:eca} is consistent with prior work showing that models encode correct information internally but fail to express it~\citep{burns2023discovering, orgad2025llms}. \citet{turpin2023unfaithful}further show that generated explanations can misstate the factors driving a prediction. What distinguishes our finding is that the correct evidence appears directly in the model's generated output: in Both-Hit (A) pairs, which average 38\% of \textit{Pred1} failures on MS-Test (26--49\% across models), the model cites the error sentence on the error note and the corresponding sentence on the correct note, yet predicts error on both. The failure is thus detectable from model outputs alone, without access to internal representations. \eca{} is deliberately a localization (evidence-citation) diagnostic and not a test of comprehension---we do not claim the model understands the sentence it cites---and the localization finding is robust to how overlap is measured: an orthogonal embedding-based Recall@1 criterion reproduces both the magnitude and the per-model ranking of TP localization (Appendix~\ref{app:robustness}, Table~\ref{tab:eca-embed}).
 
This separation between evidence production and clinical judgment suggests two directions for intervention. First, contrastive fine-tuning, which trains models to distinguish paired samples, could target the judgment component while preserving the localization ability models already demonstrate. Second, pipeline architectures that decompose the task into localization followed by judgment may better match the model's existing capability structure, in which localization is relatively strong (Both-Hit and TP-Only together account for a mean of 70\% of \textit{Pred1} pairs on MS-Test) but judgment is weak.
 
Until discriminative ability improves, the high recall of yes-biased models may support pre-filtering workflows with mandatory human review, provided the base-rate of errors is high enough that precision remains workable and alert fatigue is monitored: the model flags candidate errors at high sensitivity, and clinicians provide the judgment that the model lacks. Standalone deployment for clinical error detection may not be supported by current evidence.

\subsection{Limitations}
\label{sec:limitations}
 
All evaluations use zero-shot prompting with two prompt variants and, for sampling-based decoding, a single inference run per configuration. Few-shot or task-specific fine-tuning may improve discrimination; however, zero-shot reflects the most realistic deployment scenario. Our primary measure, the four-configuration mean, is anchored by two fully reproducible greedy-decoding runs, which mitigates single-run sampling variance. Conservative prompting shifts \bcr{} by up to 12.8~pp in either direction (Figure~\ref{fig:crossconfig}), yet at least 13 of 15 models stay below the 25\% balanced-random level under either prompt alone, so prompt-level interventions redistribute errors rather than close the discrimination gap.

Our systematic evaluation covers open-weight models up to 70B parameters, using a single proprietary reference point (GPT-5~mini on MS-Test, Section~\ref{sec:results:deception}) rather than a full proprietary sweep. This scoping reflects both computational and budget constraints and a methodological concern: proprietary models do not disclose their training data, so we cannot verify whether the benchmark source datasets appeared during their training; the open-weight models we evaluate are not fully immune to this data-leakage risk, but their training-data documentation at least allows partial verification. More importantly, the 3B--70B range is sufficient to establish that discrimination failure is pervasive and that scaling alone provides limited gains. Our contribution is diagnostic: we aim to characterize where and why pairwise discrimination fails, not to produce a ranking of models for clinical deployment.

A further scope limitation is the error type. Clinical errors also include insertions and omissions, but the public benchmarks with released pairs inject substitution-form errors, so our empirical results are confined to that form. The paired machinery extends directly once such data exist: an inserted false statement is scored as the error note against the note without it, and an omitted finding as the incomplete note against the complete one, with \eca{} localizing insertions on the present text and omissions on the complete-note side. Evaluating \bcr{} on paired insertion and omission data is the immediate next step this framework enables.

\eca{} substring matching was designed for single-sentence errors; MedRECT-JA's multi-sentence structures reduce matching granularity, and TP localization may partly reflect entity salience rather than true error localization. These factors affect the granularity of the diagnosis but do not undermine the core observation. All 15 models exceed their respective random baselines, and the localization-judgment gap is consistently observed across all four benchmarks. Our analysis also does not investigate the internal mechanisms that produce this gap; understanding why models fail to convert correct evidence into correct verdicts is an important next step that the diagnostic framework presented here should enable.

\acks{Our work was partly supported by NSF awards 2443639 and 2552481, and NIH awards, P20GM103446 and U54GM104941.}

\clearpage

\bibliography{references}

@book{iom1999err,
    title     = {To Err Is Human: Building a Safer Health System},
    editor    = {Kohn, Linda T. and Corrigan, Janet M. and Donaldson, Molla S.},               
    publisher = {National Academies Press},
    address   = {Washington, DC},
    year      = {2000},
    doi       = {10.17226/9728},
}

@inproceedings{BenAbacha2024_MEDIQA-CORR,
  title     = {Overview of the {MEDIQA}-{CORR} 2024 Shared Task on Medical Error Detection and Correction},
  author    = {Ben Abacha, Asma and Yim, Wen-wai and Fu, Yujuan and Sun, Zhaoyi and Xia, Fei and Yetisgen, Meliha},
  booktitle = {Proceedings of the 6th Clinical Natural Language Processing Workshop (ClinicalNLP 2024)},
  pages     = {596--603},
  year      = {2024},
  doi       = {10.18653/v1/2024.clinicalnlp-1.57},
}

@inproceedings{BenAbacha2025_MEDEC,
  title     = {{MEDEC}: A Benchmark for Medical Error Detection and Correction in Clinical Notes},
  author    = {Ben Abacha, Asma and Yim, Wen-wai and Fu, Yujuan and Sun, Zhaoyi and Yetisgen, Meliha and Xia, Fei and Lin, Thomas},
  booktitle = {Findings of the Association for Computational Linguistics: ACL 2025},
  pages     = {22539--22550},
  year      = {2025},
  doi       = {10.18653/v1/2025.findings-acl.1159},
}

@article{Iwase2025_MedRECT,
  title   = {{MedRECT}: A Medical Reasoning Benchmark for Error Correction in Clinical Texts},
  author  = {Iwase, Naoto and Okuyama, Hiroki and Iwasawa, Junichiro},
  journal = {arXiv preprint arXiv:2511.00421},
  year    = {2025},
}

@inproceedings{ma2026mederrbench,
  title     = {{M}ed{E}rr{B}ench: A Fine-Grained Multilingual Benchmark for Medical Error Detection and Correction with Clinical Expert Annotations},
  author    = {Ma, Congbo and Zhang, Yichun and Al-Jazzazi, Yousef and Foisal, Ahamed and Sharma, Laasya and Sadqi, Yousra and Saleh, Khaled and Mallat, Jihad and Shamout, Farah E.},
  booktitle = {Findings of the Association for Computational Linguistics: ACL 2026},
  address   = {San Diego, California, United States},
  publisher = {Association for Computational Linguistics},
  pages     = {11802--11827},
  year      = {2026},
  doi       = {10.18653/v1/2026.findings-acl.573},
}

@article{grattafiori2024llama3,
  title   = {The {Llama} 3 Herd of Models},
  author  = {Grattafiori, Aaron and Dubey, Abhimanyu and Jauhri, Abhinav and others},
  journal = {arXiv preprint arXiv:2407.21783},
  year    = {2024},
}

@article{qwen3_2025,
  title   = {Qwen3 Technical Report},
  author  = {Yang, An and Li, Anfeng and Yang, Baosong and others},
  journal = {arXiv preprint arXiv:2505.09388},
  year    = {2025},
}

@article{gemma3_2025,
  title   = {Gemma 3 Technical Report},
  author  = {{Gemma Team} and Kamath, Aishwarya and Ferret, Johan and Pathak, Shreya and others},
  journal = {arXiv preprint arXiv:2503.19786},
  year    = {2025},
}

@article{jiang2023mistral,
  title   = {Mistral 7{B}},
  author  = {Jiang, Albert Q. and Sablayrolles, Alexandre and Mensch, Arthur and others},
  journal = {arXiv preprint arXiv:2310.06825},
  year    = {2023},
}

@article{microsoft2025phi4mini,
  title   = {Phi-4-Mini Technical Report: Compact yet Powerful Multimodal Language Models via Mixture-of-LoRAs},
  author  = {{Microsoft} and Abouelenin, Abdelrahman and Ashfaq, Atabak and Atkinson, Adam and others},
  journal = {arXiv preprint arXiv:2503.01743},
  year    = {2025},
}

@article{Sellergren2025_MedGemma,
  title   = {{MedGemma} Technical Report},
  author  = {Sellergren, Andrew and Kazemzadeh, Sahar and Jaroensri, Tiam and Kiraly, Atilla and Traverse, Madeleine and others},
  journal = {arXiv preprint arXiv:2507.05201},
  year    = {2025},
}

@inproceedings{Corbeil2025_MediPhi,
  title     = {A Modular Approach for Clinical {SLM}s Driven by Synthetic Data with Pre-Instruction Tuning, Model Merging, and Clinical-Tasks Alignment},
  author    = {Corbeil, Jean-Philippe and Dada, Amin and Attendu, Jean-Michel and Ben Abacha, Asma and Sordoni, Alessandro and Caccia, Lucas and Beaulieu, Fran{\c{c}}ois and Lin, Thomas and Kleesiek, Jens and Vozila, Paul},
  booktitle = {Proceedings of the 63rd Annual Meeting of the Association for Computational Linguistics (Volume 1: Long Papers)},
  pages     = {19352--19374},
  year      = {2025},
  doi       = {10.18653/v1/2025.acl-long.950},
}

@inproceedings{zhang2024ultramedical,
  title     = {{UltraMedical}: Building Specialized Generalists in Biomedicine},
  author    = {Zhang, Kaiyan and Zeng, Sihang and Hua, Ermo and Ding, Ning and Chen, Zhang-Ren and Ma, Zhiyuan and Li, Haoxin and Cui, Ganqu and Qi, Biqing and Zhu, Xuekai and Lv, Xingtai and Hu, Jin-Fang and Liu, Zhiyuan and Zhou, Bowen},
  booktitle = {Advances in Neural Information Processing Systems},
  volume    = {37},
  pages     = {26045--26081},
  year      = {2024},
  doi       = {10.52202/079017-0819},
}

@inproceedings{kwon2023vllm,
  title     = {Efficient Memory Management for Large Language Model Serving with {PagedAttention}},
  author    = {Kwon, Woosuk and Li, Zhuohan and Zhuang, Siyuan and Sheng, Ying and Zheng, Lianmin and Yu, Cody Hao and Gonzalez, Joseph and Zhang, Hao and Stoica, Ion},
  booktitle = {Proceedings of the 29th Symposium on Operating Systems Principles},
  pages     = {611--626},
  year      = {2023},
  doi       = {10.1145/3600006.3613165},
}

@article{makary2016medical,
  title   = {Medical error---the third leading cause of death in the {US}},
  author  = {Makary, Martin A. and Daniel, Michael},
  journal = {BMJ},
  volume  = {353},
  pages   = {i2139},
  year    = {2016},
  doi     = {10.1136/bmj.i2139},
}

@inproceedings{sharma2024sycophancy,
  title     = {Towards Understanding Sycophancy in Language Models},
  author    = {Sharma, Mrinank and Tong, Meg and Korbak, Tomasz and Duvenaud, David and Askell, Amanda and Bowman, Samuel R. and Cheng, Newton and Durmus, Esin and Hatfield-Dodds, Zac and Johnston, Scott R. and Kravec, Shauna and Maxwell, Timothy and McCandlish, Sam and Ndousse, Kamal and Rausch, Oliver and Schiefer, Nicholas and Yan, Da and Zhang, Miranda and Perez, Ethan},
  booktitle = {Proceedings of the International Conference on Learning Representations (ICLR)},
  year      = {2024},
  url       = {https://openreview.net/forum?id=tvhaxkMKAn}
}

@inproceedings{gardner2020evaluating,
  title     = {Evaluating Models' Local Decision Boundaries via Contrast Sets},
  author    = {Gardner, Matt and Artzi, Yoav and Basmov, Victoria and Berant, Jonathan and Bogin, Ben and Chen, Sihao and Dasigi, Pradeep and Dua, Dheeru and Elazar, Yanai and Gottumukkala, Ananth and others},
  booktitle = {Findings of the Association for Computational Linguistics: EMNLP 2020},
  pages     = {1307--1323},
  year      = {2020},
  doi       = {10.18653/v1/2020.findings-emnlp.117},
}

@inproceedings{corbeil2024iryonlp,
  title     = {{IryoNLP} at {MEDIQA-CORR} 2024: Tackling the Medical Error Detection {\&} Correction Task on the Shoulders of Medical Agents},
  author    = {Corbeil, Jean-Philippe},
  booktitle = {Proceedings of the 6th Clinical Natural Language Processing Workshop},
  pages     = {570--580},
  year      = {2024},
  doi       = {10.18653/v1/2024.clinicalnlp-1.54},
}

@inproceedings{toma2024wanglab,
  title     = {{WangLab} at {MEDIQA-CORR} 2024: Optimized {LLM}-Based Programs for Medical Error Detection and Correction},
  author    = {Toma, Augustin and Xie, Ronald and Palayew, Steven and Lawler, Patrick and Wang, Bo},
  booktitle = {Proceedings of the 6th Clinical Natural Language Processing Workshop},
  pages     = {616--623},
  year      = {2024},
  doi       = {10.18653/v1/2024.clinicalnlp-1.59},
}

@article{Agrawal2025_evaluation,
  title     = {The evaluation illusion of large language models in medicine},
  author    = {Monica Agrawal and Irene Y. Chen and Freya Gulamali and Shalmali Joshi},
  journal   = {npj Digital Medicine},
  volume    = {8},
  number    = {1},
  pages     = {600},
  year      = {2025},
  doi       = {10.1038/s41746-025-01963-x}
}

@article{kanithi2026medic,
  title   = {{MEDIC}: Comprehensive Evaluation of Leading Indicators for {LLM} Safety and Utility in Clinical Applications},
  author  = {Kanithi, Praveenkumar and Christophe, Cl{\'e}ment and Pimentel, Marco AF and Raha, Tathagata and Munjal, Prateek and Saadi, Nada and Javed, Hamza A. and Maslenkova, Svetlana and Hayat, Nasir and Rajan, Ronnie and Khan, Shadab},
  journal = {Transactions on Machine Learning Research},
  issn    = {2835-8856},
  year    = {2026},
  url     = {https://openreview.net/forum?id=pDQe9Icwb6}
}

@article{schmidgall2024bias,
  title   = {Evaluation and mitigation of cognitive biases in medical language models},
  author  = {Schmidgall, Samuel and Harris, Carl and Essien, Ime and Olshvang, Daniel and Rahman, Tawsifur and Kim, Ji Woong and Ziaei, Rojin and Eshraghian, Jason and Abadir, Peter and Chellappa, Rama},
  journal = {npj Digital Medicine},
  volume  = {7},
  number  = {1},
  pages   = {295},
  year    = {2024},
  doi     = {10.1038/s41746-024-01283-6},
}

@book{green1966sdt,
  title     = {Signal Detection Theory and Psychophysics},
  author    = {Green, David M. and Swets, John A.},
  publisher = {Wiley},
  year      = {1966},
}

@article{warstadt2020blimp,
  title   = {{BLiMP}: The Benchmark of Linguistic Minimal Pairs for {English}},
  author  = {Warstadt, Alex and Parrish, Alicia and Liu, Haokun and Mohananey, Anhad and Peng, Wei and Wang, Sheng-Fu and Bowman, Samuel R.},
  journal = {Transactions of the Association for Computational Linguistics},
  volume  = {8},
  pages   = {377--392},
  year    = {2020},
  doi     = {10.1162/tacl_a_00321},
}

@inproceedings{burns2023discovering,
  title     = {Discovering Latent Knowledge in Language Models Without Supervision},
  author    = {Burns, Collin and Ye, Haotian and Klein, Dan and Steinhardt, Jacob},
  booktitle = {Proceedings of the International Conference on Learning Representations (ICLR)},
  year      = {2023},
  url       = {https://openreview.net/forum?id=ETKGuby0hcs}
}

@inproceedings{orgad2025llms,
  title     = {{LLM}s Know More Than They Show: On the Intrinsic Representation of {LLM} Hallucinations},
  author    = {Orgad, Hadas and Toker, Michael and Gekhman, Zorik and Reichart, Roi and Szpektor, Idan and Kotek, Hadas and Belinkov, Yonatan},
  booktitle = {The Thirteenth International Conference on Learning Representations},
  year      = {2025},
  url       = {https://openreview.net/forum?id=KRnsX5Em3W}
}

@inproceedings{turpin2023unfaithful,
  title     = {Language Models Don't Always Say What They Think: Unfaithful Explanations in Chain-of-Thought Prompting},
  author    = {Turpin, Miles and Michael, Julian and Perez, Ethan and Bowman, Samuel R.},
  booktitle = {Advances in Neural Information Processing Systems (NeurIPS)},
  volume    = {36},
  year      = {2023},
  doi       = {10.52202/075280-3275}
}

@inproceedings{deyoung2020eraser,
  title     = {{ERASER}: A Benchmark to Evaluate Rationalized {NLP} Models},
  author    = {DeYoung, Jay and Jain, Sarthak and Rajani, Nazneen Fatema and Lehman, Eric and Xiong, Caiming and Socher, Richard and Wallace, Byron C.},
  booktitle = {Proceedings of the 58th Annual Meeting of the Association for Computational Linguistics (ACL)},
  pages     = {4443--4458},
  year      = {2020},
  doi       = {10.18653/v1/2020.acl-main.408},
}

@inproceedings{lipton2014thresholding,
    author    = {Lipton, Zachary C. and Elkan, Charles and Naryanaswamy, Balakrishnan},
    title     = {Optimal Thresholding of Classifiers to Maximize {F1} Measure},
    booktitle = {Machine Learning and Knowledge Discovery in Databases -- European Conference, ECML PKDD 2014},
    editor    = {Calders, Toon and Esposito, Floriana and H{\"u}llermeier, Eyke and Meo, Rosa},
    series    = {Lecture Notes in Computer Science},
    volume    = {8725},
    pages     = {225--239},
    publisher = {Springer},
    year      = {2014},
    doi       = {10.1007/978-3-662-44851-9_15}
}

@article{reinke2024pitfalls,
    author    = {Reinke, Annika and Tizabi, Minu D. and Baumgartner, Michael and Eisenmann, Matthias and Heckmann-N{\"o}tzel, Doreen and others},
    title     = {Understanding metric-related pitfalls in image analysis validation},
    journal   = {Nature Methods},
    volume    = {21},
    number    = {2},
    pages     = {182--194},
    year      = {2024},
    doi       = {10.1038/s41592-023-02150-0}
}

@article{vancalster2025landig,
    author    = {Van Calster, Ben and Collins, Gary S. and Vickers, Andrew J. and Wynants, Laure and Kerr, Kathleen F. and Barre{\~n}ada, Lasai and Varoquaux, Ga{\"e}l and Singh, Karandeep and Moons, Karel G. M. and Hernandez-Boussard, Tina and Timmerman, Dirk and McLernon, David J. and van Smeden, Maarten and Steyerberg, Ewout W. and {Topic Group 6 of the STRATOS initiative}},
    title     = {Evaluation of performance measures in predictive artificial intelligence models to support medical decisions: overview and guidance},
    journal   = {The Lancet Digital Health},
    volume    = {7},
    number    = {12},
    pages     = {100916},
    year      = {2025},
    doi       = {10.1016/j.landig.2025.100916}
}

@inproceedings{xiao2023cpack,
    author = {Xiao, Shitao and Liu, Zheng and Zhang, Peitian and Muennighoff, Niklas and Lian, Defu and Nie, Jian-Yun},
    title = {C-Pack: Packed Resources For General Chinese Embeddings},
    year = {2024},
    doi = {10.1145/3626772.3657878},
    booktitle = {Proceedings of the 47th International ACM SIGIR Conference on Research and Development in Information Retrieval},
    pages = {641--649},
    numpages = {9}
}

@article{poulain2026bias,
  title={Bias patterns in the application of {LLMs} for clinical decision support: A comprehensive study},
  author={Poulain, Raphael and Adiba, Farzana Islam and Fayyaz, Hamed and Beheshti, Rahmatollah},
  journal={Delaware Journal of Public Health},
  volume={12},
  number={1},
  pages={54--67},
  year={2026},
  doi={10.32481/djph.2026.03.10},
  note = {arXiv:2404.15149}
}

@article{fayyaz2024scalable,
  title={Enabling scalable evaluation of bias patterns in medical {LLMs}},
  author={Fayyaz, Hamed and Poulain, Raphael and Beheshti, Rahmatollah},
  journal={arXiv preprint arXiv:2410.14763},
  year={2024}
}

@article{adiba2025scoping,
  title={Bias and fairness in medical {LLMs}: An extensive scoping review},
  author={Adiba, Farzana Islam and Zhang, Yifan and Beheshti, Rahmatollah},
  journal={OSF Preprints},
  year={2025},
  doi={10.31219/osf.io/fqejh_v1}
}

@article{deka2022improved,
  title={Improved Methods To Aid Unsupervised Evidence-Based Fact Checking For Online Health News},
  author={Deka, Pritam and Jurek-Loughrey, Anna and Deepak, P},
  journal={Journal of Data Intelligence},
  volume={3},
  number={4},
  pages={474--504},
  year={2022}
}

\appendix

\section{Pair Construction Details}
\label{app:pairs}
 
BCR evaluation requires matched (error, correct) pairs from the same clinical scenario. Because the four datasets differ in their release structure, pair construction uses dataset-specific methods.
 
\paragraph{MEDEC MS-Test (MS-Test).}
MEDEC~\citep{BenAbacha2025_MEDEC} does not provide explicit pairing between error-injected and correct notes. We recover pairs using Jaccard word-overlap similarity between each error note and all clean notes, with a threshold of $\geq 0.6$. Each error note is assigned to its highest-similarity clean note under one-to-one matching (i.e., once a clean note is paired, it is removed from the candidate pool). Of 311 error notes, 286 are successfully paired; 25 remain unmatched and are excluded from BCR computation. These unmatched notes are retained for traditional pointwise metric computation.
 
\paragraph{MedErrBench-EN (MEB-EN) and MedErrBench-CN (MEB-CN).}
Both test sets from MedErrBench~\citep{ma2026mederrbench} release samples in an alternating error--correct order. We pair consecutive samples directly: for each pair of adjacent rows, one has \texttt{Error Flag = 1} and the other has \texttt{Error Flag = 0}. This yields 104 pairs for MedErrBench-EN and 100 pairs for MedErrBench-CN, with no unmatched samples.
 
\paragraph{MedRECT-JA (MRT-JA).}
MedRECT~\citep{Iwase2025_MedRECT} provides a clinical scenario ID for each note. We group notes by scenario ID and pair each error note with clean notes from the same scenario, filtering to pairs with text similarity above 0.85 (measured by character-level overlap). This yields 190 pairs. However, because MedRECT contains 190 error notes but only 105 unique clean notes, some clean notes appear in multiple pairs (many-to-one structure). If a shared clean note is systematically misclassified, all its associated pairs fail simultaneously, which may inflate the apparent degree of within-pair error correlation for this dataset.

\FloatBarrier

\section{An Upper Bound on BCR}
\label{app:bound}

We show that on matched pairs \bcr{} is bounded above by the smaller of the model's marginal sensitivity and specificity, and draw the corollary that a class-biased predictor can attain a moderate F1. In contrast, its \bcr{} is pinned at zero. This makes the F1/\bcr{} divergence of Section~\ref{sec:results:deception} a structural property rather than a feature of the particular models tested.

\paragraph{Setup.} Let the evaluation set consist of $N$ matched pairs $(x_e, x_c)$, each with one error note ($y_e = 1$) and one correct note ($y_c = 0$). Write the model's per-note predictions as $\hat{y}_e, \hat{y}_c \in \{0,1\}$. Because the pairs are matched, the $N$ error notes are exactly the positive class and the $N$ correct notes exactly the negative class, so
\[
\mathrm{sensitivity} = \frac{1}{N}\sum_{i=1}^{N} \mathbf{1}[\hat{y}_{e,i} = 1], \qquad
\mathrm{specificity} = \frac{1}{N}\sum_{i=1}^{N} \mathbf{1}[\hat{y}_{c,i} = 0],
\]
while $\mathrm{BCR} = \frac{1}{N}\sum_{i} \mathbf{1}[\hat{y}_{e,i} = 1 \wedge \hat{y}_{c,i} = 0]$ (Equation~\ref{eq:bcr}).

\paragraph{Bound.} For every pair $i$, the Both-Correct indicator is the product of the two single-note success indicators, and is therefore at most each factor:
\[
\mathbf{1}[\hat{y}_{e,i}=1 \wedge \hat{y}_{c,i}=0]
= \mathbf{1}[\hat{y}_{e,i}=1]\,\mathbf{1}[\hat{y}_{c,i}=0]
\leq \mathbf{1}[\hat{y}_{e,i}=1],
\]
and likewise $\leq \mathbf{1}[\hat{y}_{c,i}=0]$. Averaging over the $N$ pairs gives $\mathrm{BCR} \leq \mathrm{sensitivity}$ and $\mathrm{BCR} \leq \mathrm{specificity}$, hence
\[
\mathrm{BCR} \leq \min(\mathrm{sensitivity}, \mathrm{specificity}),
\]
which is Equation~\ref{eq:bcr-bound}. Equivalently, the Both-Correct pairs are the intersection of the correctly flagged error notes and the correctly cleared correct notes, and an intersection cannot exceed either set.

\paragraph{Corollary: class bias caps BCR.} The bound is governed by the weaker marginal, which a class-biased model drives toward zero. The constant ``always-error'' predictor ($\hat{y} = 1$ for every note) has $\mathrm{sensitivity} = 1$ and $\mathrm{specificity} = 0$, so $\min(\mathrm{sensitivity}, \mathrm{specificity}) = 0$ and therefore $\mathrm{BCR} = 0$: it fails every pair on the correct member. Yet on a class-balanced set its F1 is bounded well away from zero---with precision $=1/2$ and recall $=1$,
\[
\mathrm{F1} = \frac{2 \cdot \tfrac{1}{2} \cdot 1}{\tfrac{1}{2} + 1} = \frac{2}{3} \approx 0.67 .
\]
More generally, any predictor whose bias pushes one marginal toward its extreme incurs a \bcr{} ceiling at the opposite, weak marginal, however high F1 climbs through recall on the favored class. The F1/\bcr{} divergence documented in Section~\ref{sec:results:deception} thus follows from Equation~\ref{eq:bcr-bound} for any model set, and is not an artifact of the particular models we evaluate.

\FloatBarrier

\section{Evaluated Models: References and Identifiers}
\label{app:models}

Table~\ref{tab:models:apx} lists, for each of the 15 evaluated models, the technical report/publication citation, the HuggingFace repository identifier, the model family, whether it is medical-domain, and the loading precision. Midrule separators follow the three size tiers (Small 3--8B, Large 27--32B, Very Large 70B). 

\begin{table}[h] \caption{References, HuggingFace identifiers, family, domain, and loading precision for all 15 evaluated models. All models are instruction-tuned.} 
\label{tab:models:apx} 
\centering\footnotesize\setlength{\tabcolsep}{3pt}
\resizebox{\textwidth}{!}{%
\begin{tabular}{lllcc}
\toprule 
Model (Reference) & HuggingFace ID & Family & Domain & Prec. \\ 
\midrule 
Llama 3.2-3B~\citep{grattafiori2024llama3}   & \texttt{meta-llama/Llama-3.2-3B-Instruct} & Llama & General & bf16   \\ 
Gemma 3-4B~\citep{gemma3_2025}             & \texttt{google/gemma-3-4b-it} & Gemma & General & bf16               \\ 
Qwen 3-4B~\citep{qwen3_2025}              & \texttt{Qwen/Qwen3-4B-Instruct-2507} & Qwen & General & bf16        \\ 
MedGemma 4B~\citep{Sellergren2025_MedGemma} & \texttt{google/medgemma-4b-it} & Gemma & \textbf{Medical} & bf16              \\ 
MediPhi 4B~\citep{Corbeil2025_MediPhi}     & \texttt{microsoft/MediPhi-Instruct} & Phi & \textbf{Medical} & bf16         \\ 
Phi-4-mini~\citep{microsoft2025phi4mini}   & \texttt{microsoft/Phi-4-mini-instruct} & Phi & General & bf16      \\ 
Mistral 7B v0.3~\citep{jiang2023mistral}        & \texttt{mistralai/Mistral-7B-Instruct-v0.3} & Mistral & General & bf16 \\ 
Llama 3.1-8B~\citep{grattafiori2024llama3}   & \texttt{meta-llama/Llama-3.1-8B-Instruct} & Llama & General & bf16   \\ 
UltraMedical 8B~\citep{zhang2024ultramedical}   & \texttt{TsinghuaC3I/Llama-3-8B-UltraMedical} & Llama & \textbf{Medical} & bf16 \\ 
Qwen 3-8B~\citep{qwen3_2025}              & \texttt{Qwen/Qwen3-8B} & Qwen & General & bf16 \\ 
\midrule 
Gemma 3-27B~\citep{gemma3_2025}             & \texttt{google/gemma-3-27b-it} & Gemma & General & bf16              \\ 
MedGemma 27B~\citep{Sellergren2025_MedGemma} & \texttt{google/medgemma-27b-text-it} & Gemma & \textbf{Medical} & bf16        \\ 
Qwen 3-32B~\citep{qwen3_2025}              & \texttt{Qwen/Qwen3-32B} & Qwen & General & bf16 \\ 
\midrule 
Llama 3.1-70B~\citep{grattafiori2024llama3}   & \texttt{meta-llama/Llama-3.1-70B-Instruct} & Llama & General & fp8  \\ 
UltraMedical 70B~\citep{zhang2024ultramedical}   & 
\texttt{TsinghuaC3I/Llama-3-70B-UltraMedical} & Llama & \textbf{Medical} & fp8 \\
\bottomrule
\end{tabular}}
\end{table}

\FloatBarrier

\section{Prompt Templates and Decoding Settings}
\label{app:prompts}

\paragraph{Decoding settings.} Greedy decoding uses temperature $T=0$. Stochastic sampling uses $T=0.7$ and top-$p=0.9$, common defaults for open-ended generation. Each model is run once per (prompt~$\times$~decoding) configuration, yielding four runs per model--dataset pair.

\paragraph{Output format.} All prompts require a structured four-line output format with \texttt{Evidence} (a quoted text span from the note, or NA), \texttt{Analysis} (3--5 sentences of reasoning), \texttt{Confidence} (0--100), and \texttt{Error: Yes/No}. The key structural difference between neutral and conservative variants is in the system prompt and error-flagging threshold.

\paragraph{MEDEC MS-Test --- Neutral.}
\textit{System:} You are a skilled medical doctor reviewing the clinical text.

\begin{quote}\small
The following is a medical narrative about a patient. The text is either correct or contains one error.

An ``error'' must meet BOTH:
(1) The note explicitly states something incorrect/unsafe (diagnosis, management, medication, treatment, or causal organism), AND
(2) You can quote the exact text span that is wrong.

If you cannot quote an exact span, output Error: No.

Output EXACTLY 4 lines (nothing else):\\
Evidence: $\langle$copy a short exact quote from the note that is wrong, or NA$\rangle$\\
Analysis: [Your short reasoning in 3--5 sentences]\\
Confidence: [0--100]\\
Error: [No or Yes]

Clinical Note:  \{medical\_note\} 
\end{quote}

\paragraph{MEDEC MS-Test --- Conservative.}
\textit{System:} You are a conservative medical reviewer. Only flag a significant medical error if the text itself clearly supports it. Do NOT assume missing information. Do NOT invent guidelines or facts not stated in the note. If the note is incomplete or ambiguous, prefer ``no error'' with lower confidence.

The user prompt follows the same structure as neutral, with ``significant medical error'' replacing ``error.''

\paragraph{MedErrBench-EN.}
Identical to MEDEC MS-Test prompts, except the error-type list is expanded to 10 categories: diagnosis, management, treatment, pharmacotherapy, causal organism, lab/serum value interpretation, physiology, histology, anatomy, and epidemiology.

\paragraph{MedErrBench-CN.}
The same prompt structure was translated into Chinese, along with lists of Chinese-language error types. Output format keys (\texttt{Evidence}, \texttt{Analysis}, \texttt{Confidence}, \texttt{Error}) remain in English for parsing consistency.

\paragraph{MedRECT-JA.}
The same prompt structure was translated into Japanese, with nine Japanese-language error types derived from the dataset's annotation taxonomy. Output format keys remain in English.

\FloatBarrier

\section{Robustness Analyses}
\label{app:robustness}

\paragraph{Parse failure handling.}
Nine of 240 runs exceed a 5\% parse failure rate (Table~\ref{tab:parsefail}). Parse failures concentrate in UltraMedical (8B and 70B), primarily on the non-English datasets, and in Llama~3.2-3B on MedErrBench-CN. Comparing the skip strategy against random assignment yields deltas below 0.5~pp for all 15 models (Table~\ref{tab:skipvsrand}). The same 13 of 15 models fall below 25\% \bcr{} under both strategies. All \bcr{} values reported in this paper use the skip strategy; random assignment serves only as a robustness check.

\begin{table}[h]
\caption{Runs exceeding 5\% parse failure rate (9 of 240).}
\label{tab:parsefail}
\centering\small\setlength{\tabcolsep}{4pt}
\begin{tabular}{llcr}
\toprule
Model & Dataset & Config & parse-failure rate (\%) \\
\midrule
UltraMed-8B  & MEB-CN & neut-gre & 24.5 \\
UltraMed-70B & MRT-JA & cons-sam & 22.0 \\
UltraMed-70B & MRT-JA & neut-sam & 14.9 \\
Llama 3.2-3B & MEB-CN & cons-sam &  9.5 \\
UltraMed-8B  & MEB-CN & neut-sam &  9.5 \\
Llama 3.2-3B & MEB-CN & neut-sam &  9.0 \\
UltraMed-8B  & MRT-JA & neut-sam &  8.8 \\
UltraMed-70B & MRT-JA & cons-gre &  7.5 \\
UltraMed-8B  & MS-Test & neut-sam &  7.4 \\
\bottomrule
\end{tabular}
\end{table}

\begin{table}[h]
\caption{Skip vs.\ random-assignment \bcr{} comparison (four-configuration mean, cross-dataset mean). All deltas $<$ 0.5~pp.}
\label{tab:skipvsrand}
\centering\small\setlength{\tabcolsep}{4pt}
\begin{tabular}{lrrr}
\toprule
Model & Skip (\%) & Random (\%) & $\Delta$ \\
\midrule
Qwen 3-32B       & 28.0 & 28.0 &  0.0 \\
UltraMed-70B     & 25.7 & 26.1 & $-$0.4 \\
MedGemma 27B     & 23.1 & 23.1 &  0.0 \\
Llama 3.1-70B    & 20.0 & 20.1 & $-$0.1 \\
UltraMed-8B      & 20.0 & 20.3 & $-$0.3 \\
Qwen 3-4B        & 17.0 & 17.0 &  0.0 \\
MedGemma 4B      & 15.0 & 15.0 &  0.0 \\
Llama 3.1-8B     & 14.6 & 14.7 & $-$0.1 \\
Phi-4-mini       & 12.7 & 12.6 & $+$0.1 \\
Qwen 3-8B        & 12.1 & 12.1 &  0.0 \\
MediPhi 4B       & 10.7 & 10.7 &  0.0 \\
Gemma 3-27B      & 10.2 & 10.2 &  0.0 \\
Llama 3.2-3B     & 10.2 & 10.5 & $-$0.3 \\
Mistral 7B       &  7.1 &  7.3 & $-$0.2 \\
Gemma 3-4B       &  4.6 &  4.6 &  0.0 \\
\bottomrule
\end{tabular}
\end{table}

\FloatBarrier

\paragraph{Threshold sensitivity.} We checked the ECA word-coverage threshold at 0.5, 0.6, and 0.7, and the MS-Test Jaccard pairing threshold at 0.5, 0.6, and 0.7. Per-model TP-localization rates shift by $<$ 3~pp, and the rank ordering of models is preserved. The MRT-JA character-overlap threshold of 0.85 yields the full 190-pair set; lowering to 0.75 adds no pairs because the same-scenario notes already exceed 0.85.

\paragraph{Embedding-based localization robustness.} \eca{}'s TP localization is scored by an absolute substring/word-coverage rule, which raises the concern that the localization finding could be an artifact of that particular criterion or of incidental word overlap. To test this, we re-scored TP localization on the \emph{same} model outputs (no re-inference) with a methodologically orthogonal, embedding-based criterion. For each \textit{Pred1} pair on MS-Test, we embed the cited \texttt{Evidence} span and every sentence of the error note with a sentence encoder, rank sentences by cosine similarity, and count a hit only when the gold error sentence is the single most similar sentence (\emph{Recall@1})---a relative-ranking test that is stricter than an absolute threshold and not tied to any word-overlap cutoff. We use two encoders, one biomedical (S-PubMedBert~\citep{deka2022improved}, checkpoint \texttt{pritamdeka/S-PubMedBert-MS-MARCO}) and one general (BGE-large~\citep{xiao2023cpack}, checkpoint \texttt{BAAI/bge-large-en-v1.5}), and pool over the four prompt--decoding configurations of all 15 models, giving $N = 9{,}428$ localization judgments. As shown in Table~\ref{tab:eca-embed}, pooled Recall@1 is 65.3\% / 67.1\% against a 12.6\% random baseline, all 15 models clear the baseline under both encoders, and per-model Recall@1 correlates with the substring TP-localization rate at Pearson $r = 0.99$ / $0.98$; the pooled substring rate (69.4\%) reproduces the 69\% reported in Section~\ref{sec:results:eca}. A methodologically different localization criterion therefore reproduces both the per-model ranking and the magnitude of the finding. We present this as robustness of the localization signal to how it is measured, not as a claim of semantic abstraction from word overlap: the agreement reflects that models predominantly cite the error sentence verbatim, which, if anything, sharpens the localization--judgment gap, since the model reproduces the erroneous sentence in its own evidence yet still labels both the erroneous and the corrected note as containing an error.

\begin{table}[h]
\caption{Embedding-based re-scoring of TP localization on MS-Test \textit{Pred1} pairs ($N = 9{,}428$ judgments pooled over 15 models $\times$ 4 configurations). Recall@1 counts a hit only when the gold error sentence is the single most similar sentence to the cited evidence. Random baseline $= 12.6\%$. Pearson $r$ is the per-model correlation with the substring TP-localization rate.}
\label{tab:eca-embed}
\centering\small\setlength{\tabcolsep}{5pt}
\begin{tabular}{lccccc}
\toprule
Encoder & Recall@1 & Recall@3 & MRR & Pearson $r$ & Above baseline \\
\midrule
S-PubMedBert (biomedical) & 65.3\% & 73.8\% & 0.73 & 0.99 & 15/15 \\
BGE-large (general)       & 67.1\% & 82.5\% & 0.77 & 0.98 & 15/15 \\
\bottomrule
\end{tabular}
\end{table}

\begin{figure}[h]
\centering
\includegraphics[width=0.88\textwidth]{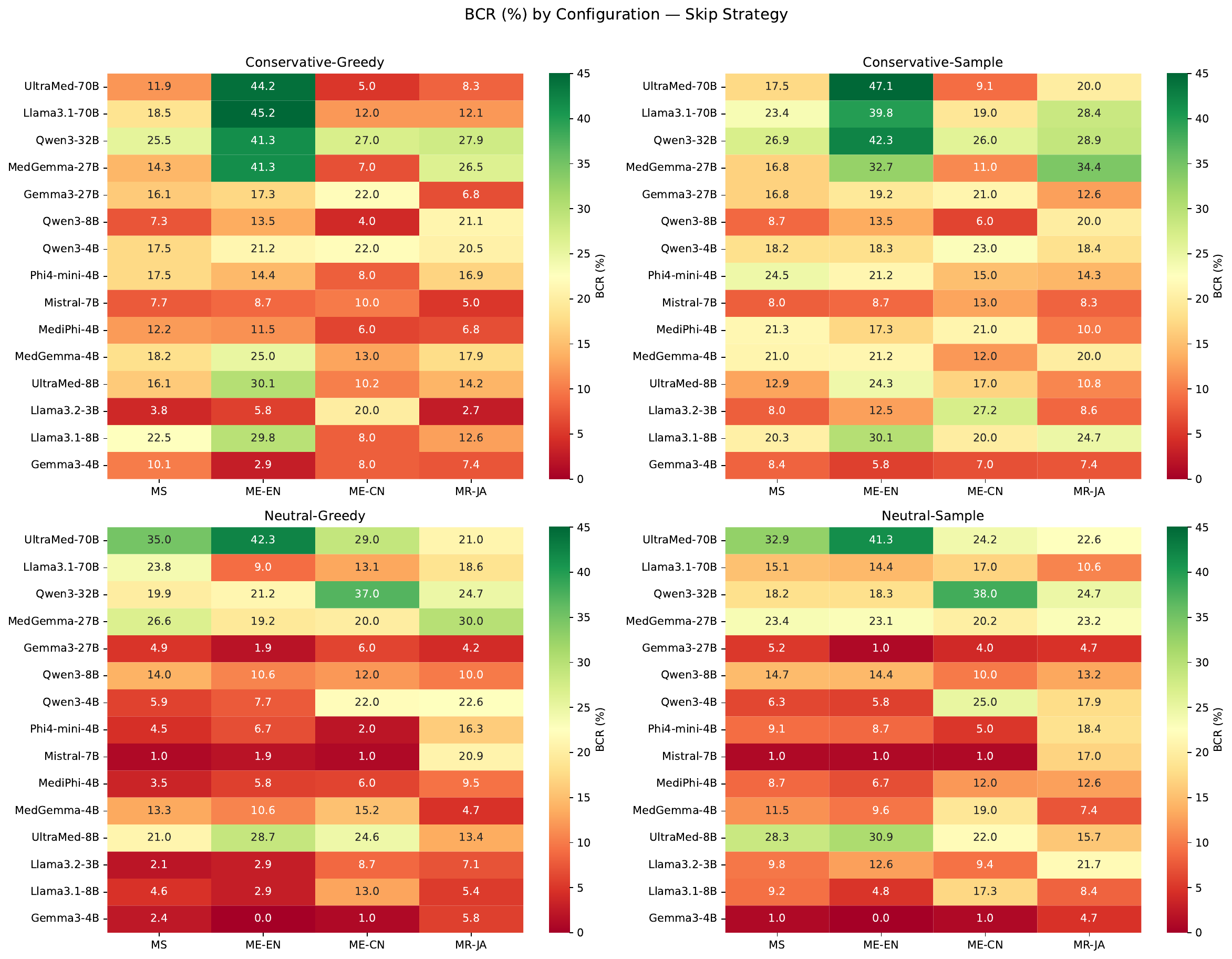}
\caption{\bcr{} (\%) by configuration. Rank ordering is largely stable; absolute values vary up to 36.2~pp.}
\label{fig:crossconfig}
\end{figure}

\paragraph{Cross-configuration \bcr{}.}
Figure~\ref{fig:crossconfig} shows \bcr{} for all 15 models under each of the four configurations. The rank ordering of models is largely stable across configurations, but absolute \bcr{} values vary substantially (up to 36.2~pp for Llama~3.1-70B).

\FloatBarrier

\section{Prediction Bias Mediation Analysis}
\label{app:mediation}

Section~\ref{sec:results:deception} reports that prediction bias (error-flag rate) mediates the relationship between F1 and \bcr{}. Tables~\ref{tab:mediation}--\ref{tab:independence} provide per-dataset details.

\begin{table}[h]
\caption{Per-dataset error-flag-rate mediation correlations (Pearson $r$, $n = 15$ models per dataset). Error-flag rate suppresses \bcr{} across all datasets and strongly inflates F1 on three of four (weaker on MEB-EN); the direct F1--\bcr{} correlation varies in sign across datasets and is near zero only when pooled.}
\label{tab:mediation}
\centering\small\setlength{\tabcolsep}{5pt}
\begin{tabular}{lccc}
\toprule
Dataset & Error-flag $\to$ F1 & Error-flag $\to$ \bcr{} & F1 $\to$ \bcr{} \\
\midrule
MS-Test  & $+$0.946 & $-$0.855 & $-$0.659 \\
MEB-EN  & $+$0.406 & $-$0.737 & $+$0.310 \\
MEB-CN  & $+$0.911 & $-$0.530 & $-$0.142 \\
MRT-JA     & $+$0.963 & $-$0.480 & $-$0.233 \\
\midrule
All ($n = 60$) & $+$0.853 & $-$0.488 & $-$0.058 \\
\bottomrule
\end{tabular}
\end{table}

\begin{table}[h]
\caption{Top-3 models by F1 vs.\ top-3 by \bcr{} (4-config mean). On 3
  of 4 datasets, no model appears in both top-3 lists.}
\label{tab:top3}
\centering\footnotesize\setlength{\tabcolsep}{3pt}
\resizebox{\textwidth}{!}{%
\begin{tabular}{llll}
\toprule
Dataset & Top-3 by F1 & Top-3 by \bcr{} & Overlap \\
\midrule
MS-Test     & Gemma 3-4B, Llama 3.2-3B, Mistral 7B 
            & UltraMed-70B, Qwen 3-32B, MedGemma 27B & 0/3 \\
MEB-EN       & Qwen 3-32B, UltraMed-70B, Llama 3.1-70B
            & UltraMed-70B, Qwen 3-32B, MedGemma 27B  & 2/3 \\
MEB-CN        & Mistral 7B, Gemma 3-4B, MediPhi 4B
            & Qwen 3-32B, Qwen 3-4B, UltraMed-8B      & 0/3 \\
MRT-JA       & Gemma 3-27B, Gemma 3-4B, Llama 3.2-3B
            & MedGemma 27B, Qwen 3-32B, Qwen 3-4B     & 0/3 \\
\bottomrule
\end{tabular}}
\end{table}

\begin{table}[h]
\caption{Per-dataset independence ratios. \emph{Ratio} is the mean of per-model independence ratios $R_{\text{independence}}$ (Equation~\ref{eq:bcr-ratio}), not the quotient of the two column means. Actual \bcr{} falls below the expected value (sensitivity $\times$ specificity) for all 60 model--dataset entries.}
\label{tab:independence}
\centering\small\setlength{\tabcolsep}{5pt}
\begin{tabular}{lcccc}
\toprule
Dataset & Expected \bcr{} & Actual \bcr{} & Ratio & Below \\
\midrule
MS-Test        & 21.9\% & 14.2\% & 0.685$\times$ & 15/15 \\
MEB-EN  & 21.7\% & 17.8\% & 0.844$\times$ & 15/15 \\
MEB-CN  & 23.5\% & 14.4\% & 0.621$\times$ & 15/15 \\
MRT-JA     & 20.3\% & 15.2\% & 0.765$\times$ & 15/15 \\
\midrule
All            & 21.9\% & 15.4\% & 0.729$\times$ & 60/60 \\
\bottomrule
\end{tabular}
\end{table}

\FloatBarrier

\section{Detailed Results}
\label{app:detailed}

\subsection{Full Traditional Metrics (All 240 Runs)} \label{app:trad240} 
Tables~\ref{tab:trad240:cg}--\ref{tab:trad240:ns} report per-run traditional metrics for every inference run in the study, organized by configuration. Each table contains 60 rows (15 models $\times$ 4 datasets), sorted by balanced accuracy within each dataset. Flag\% (error-flag rate) is the fraction of notes flagged as containing an error.

{\footnotesize
\setlength{\tabcolsep}{3pt}
\renewcommand{\arraystretch}{0.95}

\begin{longtable}{llrrrrrrr}
\caption{Traditional metrics --- Conservative-Greedy configuration (all 15 models $\times$ 4 datasets = 60 runs). Sorted by balanced accuracy within each dataset. Flag\%(error-flag rate) is the fraction of notes flagged as containing an error.}\label{tab:trad240:cg} \\
\toprule
Dataset & Model & BalAcc & F1 & Prec & Rec & Spec & MCC & Flag\% \\
\midrule
\endfirsthead
\multicolumn{9}{l}{\textit{(Table \thetable\ continued from previous page)}} \\
\toprule
Dataset & Model & BalAcc & F1 & Prec & Rec & Spec & MCC & Flag\% \\
\midrule
\endhead
\midrule
\multicolumn{9}{r}{\textit{(continued on next page)}} \\
\endfoot
\bottomrule
\endlastfoot
MS-Test         & Qwen 3-32B     & .584 & .564 & .616 & .521 & .647 & .169 & 44.1 \\
                & Llama 3.1-70B  & .566 & .375 & .690 & .257 & .874 & .166 & 19.4 \\
                & MedGemma 27B   & .555 & .300 & .720 & .190 & .920 & .159 & 13.7 \\
                & UltraMed-70B   & .554 & .262 & .778 & .158 & .951 & .177 & 10.6 \\
                & Llama 3.1-8B   & .550 & .539 & .575 & .506 & .593 & .100 & 45.9 \\
                & MedGemma 4B    & .549 & .482 & .588 & .408 & .689 & .101 & 36.2 \\
                & Gemma 3-27B    & .546 & .513 & .576 & .463 & .629 & .094 & 41.9 \\
                & Qwen 3-4B      & .531 & .620 & .544 & .720 & .343 & .068 & 69.0 \\
                & UltraMed-8B    & .527 & .368 & .575 & .270 & .783 & .062 & 24.5 \\
                & Gemma 3-4B     & .525 & .675 & .535 & .913 & .136 & .079 & 88.9 \\
                & MediPhi 4B     & .520 & .657 & .533 & .855 & .185 & .055 & 83.6 \\
                & Phi-4-mini     & .517 & .604 & .533 & .698 & .336 & .036 & 68.2 \\
                & Qwen 3-8B      & .511 & .238 & .560 & .151 & .871 & .031 & 14.1 \\
                & Llama 3.2-3B   & .505 & .680 & .523 & .971 & .038 & .026 & 96.6 \\
                & Mistral 7B     & .496 & .658 & .519 & .900 & .091 & $-$.015 & 90.5 \\
\midrule
MedErrBench-EN  & UltraMed-70B   & .692 & .640 & .770 & .548 & .837 & .402 & 35.6 \\
                & Llama 3.1-70B  & .688 & .709 & .664 & .760 & .615 & .379 & 57.2 \\
                & MedGemma 27B   & .683 & .612 & .788 & .500 & .865 & .393 & 31.7 \\
                & Qwen 3-32B     & .673 & .721 & .629 & .846 & .500 & .369 & 67.3 \\
                & UltraMed-8B    & .594 & .567 & .604 & .534 & .654 & .189 & 44.0 \\
                & Qwen 3-4B      & .582 & .690 & .548 & .933 & .231 & .229 & 85.1 \\
                & Llama 3.1-8B   & .572 & .621 & .557 & .702 & .442 & .149 & 63.0 \\
                & MedGemma 4B    & .558 & .521 & .568 & .481 & .635 & .117 & 42.3 \\
                & Gemma 3-27B    & .553 & .669 & .531 & .904 & .202 & .148 & 85.1 \\
                & Phi-4-mini     & .538 & .657 & .523 & .885 & .192 & .107 & 84.6 \\
                & MediPhi 4B     & .538 & .671 & .521 & .942 & .135 & .130 & 90.4 \\
                & Qwen 3-8B      & .529 & .395 & .552 & .308 & .750 & .064 & 27.9 \\
                & Llama 3.2-3B   & .505 & .658 & .503 & .952 & .058 & .021 & 94.7 \\
                & Mistral 7B     & .505 & .648 & .503 & .913 & .096 & .017 & 90.9 \\
                & Gemma 3-4B     & .490 & .649 & .495 & .942 & .038 & $-$.045 & 95.2 \\
\midrule
MedErrBench-CN  & Qwen 3-32B     & .590 & .474 & .661 & .370 & .810 & .200 & 28.0 \\
                & Qwen 3-4B      & .575 & .573 & .576 & .570 & .580 & .150 & 49.5 \\
                & Gemma 3-27B    & .570 & .623 & .555 & .710 & .430 & .146 & 64.0 \\
                & Llama 3.1-70B  & .545 & .222 & .765 & .130 & .960 & .161 & 8.5 \\
                & MedGemma 4B    & .525 & .371 & .549 & .280 & .770 & .057 & 25.5 \\
                & UltraMed-70B   & .525 & .112 & .857 & .060 & .990 & .136 & 3.5 \\
                & MedGemma 27B   & .525 & .188 & .647 & .110 & .940 & .090 & 8.5 \\
                & Llama 3.2-3B   & .524 & .456 & .534 & .398 & .649 & .049 & 37.4 \\
                & Gemma 3-4B     & .520 & .631 & .512 & .820 & .220 & .050 & 80.0 \\
                & Qwen 3-8B      & .515 & .110 & .667 & .060 & .970 & .072 & 4.5 \\
                & Llama 3.1-8B   & .510 & .246 & .533 & .160 & .860 & .028 & 15.0 \\
                & UltraMed-8B    & .507 & .236 & .517 & .153 & .860 & .018 & 14.6 \\
                & Mistral 7B     & .505 & .643 & .503 & .890 & .120 & .016 & 88.5 \\
                & MediPhi 4B     & .495 & .625 & .497 & .840 & .150 & $-$.014 & 84.5 \\
                & Phi-4-mini     & .495 & .625 & .497 & .840 & .150 & $-$.014 & 84.5 \\
\midrule
MedRECT-JA      & Qwen 3-32B     & .585 & .516 & .763 & .389 & .781 & .174 & 32.9 \\
                & Qwen 3-4B      & .576 & .741 & .691 & .800 & .352 & .168 & 74.6 \\
                & Llama 3.1-70B  & .558 & .346 & .792 & .221 & .895 & .145 & 18.0 \\
                & MedGemma 27B   & .551 & .638 & .685 & .598 & .505 & .099 & 56.1 \\
                & Qwen 3-8B      & .545 & .572 & .689 & .489 & .600 & .086 & 45.8 \\
                & MedGemma 4B    & .534 & .711 & .665 & .763 & .305 & .074 & 73.9 \\
                & UltraMed-8B    & .525 & .441 & .681 & .326 & .724 & .052 & 30.8 \\
                & Phi-4-mini     & .516 & .734 & .652 & .841 & .190 & .040 & 83.0 \\
                & Gemma 3-27B    & .512 & .755 & .650 & .900 & .124 & .037 & 89.2 \\
                & UltraMed-70B   & .508 & .213 & .667 & .126 & .889 & .023 & 12.1 \\
                & Gemma 3-4B     & .507 & .753 & .648 & .900 & .114 & .022 & 89.5 \\
                & MediPhi 4B     & .507 & .749 & .648 & .889 & .124 & .020 & 88.5 \\
                & Llama 3.2-3B   & .496 & .768 & .639 & .962 & .029 & $-$.023 & 96.6 \\
                & Llama 3.1-8B   & .495 & .708 & .641 & .789 & .200 & $-$.012 & 79.3 \\
                & Mistral 7B     & .458 & .688 & .622 & .769 & .147 & $-$.100 & 79.9 \\
\end{longtable}

\begin{longtable}{llrrrrrrr}
\caption{Traditional metrics --- Conservative-Sampling configuration (all 15 models $\times$ 4 datasets = 60 runs). Sorted by balanced accuracy within each dataset. Flag\%(error-flag rate) is the fraction of notes flagged as containing an error.}\label{tab:trad240:cs} \\
\toprule
Dataset & Model & BalAcc & F1 & Prec & Rec & Spec & MCC & Flag\% \\
\midrule
\endfirsthead
\multicolumn{9}{l}{\textit{(Table \thetable\ continued from previous page)}} \\
\toprule
Dataset & Model & BalAcc & F1 & Prec & Rec & Spec & MCC & Flag\% \\
\midrule
\endhead
\midrule
\multicolumn{9}{r}{\textit{(continued on next page)}} \\
\endfoot
\bottomrule
\endlastfoot
MS-Test         & Llama 3.1-70B  & .588 & .462 & .688 & .347 & .829 & .200 & 26.3 \\
                & UltraMed-70B   & .570 & .355 & .730 & .235 & .906 & .188 & 16.8 \\
                & MedGemma 27B   & .569 & .345 & .737 & .225 & .913 & .188 & 15.9 \\
                & Qwen 3-32B     & .566 & .543 & .596 & .498 & .633 & .132 & 43.6 \\
                & Gemma 3-27B    & .555 & .518 & .588 & .463 & .647 & .112 & 41.0 \\
                & Phi-4-mini     & .542 & .606 & .555 & .669 & .416 & .088 & 62.8 \\
                & MedGemma 4B    & .542 & .476 & .578 & .405 & .678 & .087 & 36.5 \\
                & Qwen 3-4B      & .530 & .616 & .543 & .711 & .350 & .065 & 68.2 \\
                & Gemma 3-4B     & .523 & .676 & .533 & .923 & .122 & .076 & 90.1 \\
                & MediPhi 4B     & .521 & .599 & .537 & .678 & .364 & .044 & 65.8 \\
                & Llama 3.1-8B   & .517 & .446 & .544 & .377 & .657 & .036 & 36.1 \\
                & UltraMed-8B    & .505 & .341 & .527 & .252 & .758 & .011 & 24.7 \\
                & Qwen 3-8B      & .501 & .227 & .523 & .145 & .857 & .002 & 14.4 \\
                & Llama 3.2-3B   & .498 & .656 & .520 & .891 & .105 & $-$.007 & 89.3 \\
                & Mistral 7B     & .493 & .654 & .517 & .891 & .094 & $-$.025 & 89.8 \\
\midrule
MedErrBench-EN  & UltraMed-70B   & .692 & .680 & .708 & .654 & .731 & .386 & 46.2 \\
                & Qwen 3-32B     & .692 & .742 & .639 & .885 & .500 & .417 & 69.2 \\
                & Llama 3.1-70B  & .666 & .718 & .624 & .846 & .485 & .356 & 68.1 \\
                & MedGemma 27B   & .644 & .575 & .714 & .481 & .808 & .305 & 33.7 \\
                & Gemma 3-27B    & .572 & .683 & .542 & .923 & .221 & .202 & 85.1 \\
                & Phi-4-mini     & .572 & .674 & .544 & .885 & .260 & .185 & 81.2 \\
                & Llama 3.1-8B   & .570 & .594 & .565 & .625 & .515 & .140 & 55.6 \\
                & Qwen 3-4B      & .562 & .681 & .536 & .933 & .192 & .186 & 87.0 \\
                & UltraMed-8B    & .561 & .486 & .589 & .413 & .709 & .128 & 35.3 \\
                & MedGemma 4B    & .529 & .500 & .533 & .471 & .587 & .058 & 44.2 \\
                & Mistral 7B     & .529 & .673 & .515 & .971 & .087 & .124 & 94.2 \\
                & MediPhi 4B     & .519 & .618 & .513 & .779 & .260 & .045 & 76.0 \\
                & Gemma 3-4B     & .514 & .667 & .508 & .971 & .058 & .071 & 95.7 \\
                & Qwen 3-8B      & .514 & .380 & .525 & .298 & .731 & .032 & 28.4 \\
                & Llama 3.2-3B   & .510 & .641 & .506 & .875 & .144 & .028 & 86.5 \\
\midrule
MedErrBench-CN  & Qwen 3-32B     & .600 & .487 & .679 & .380 & .820 & .223 & 28.0 \\
                & Qwen 3-4B      & .580 & .584 & .578 & .590 & .570 & .160 & 51.0 \\
                & Llama 3.1-70B  & .570 & .434 & .635 & .330 & .810 & .160 & 26.0 \\
                & UltraMed-8B    & .565 & .351 & .697 & .235 & .896 & .174 & 17.0 \\
                & Gemma 3-27B    & .555 & .608 & .543 & .690 & .420 & .114 & 63.5 \\
                & Llama 3.1-8B   & .555 & .378 & .628 & .270 & .840 & .134 & 21.5 \\
                & UltraMed-70B   & .546 & .182 & .909 & .101 & .990 & .199 & 5.5 \\
                & MedGemma 27B   & .545 & .260 & .696 & .160 & .930 & .141 & 11.5 \\
                & Llama 3.2-3B   & .542 & .520 & .556 & .489 & .596 & .085 & 44.8 \\
                & Qwen 3-8B      & .530 & .145 & .800 & .080 & .980 & .138 & 5.0 \\
                & MediPhi 4B     & .530 & .618 & .521 & .760 & .300 & .068 & 73.0 \\
                & MedGemma 4B    & .515 & .312 & .537 & .220 & .810 & .037 & 20.5 \\
                & Mistral 7B     & .515 & .639 & .509 & .860 & .170 & .041 & 84.5 \\
                & Phi-4-mini     & .510 & .623 & .506 & .810 & .210 & .025 & 80.0 \\
                & Gemma 3-4B     & .485 & .593 & .490 & .750 & .220 & $-$.035 & 76.5 \\
\midrule
MedRECT-JA      & MedGemma 27B   & .627 & .700 & .749 & .658 & .596 & .245 & 56.8 \\
                & Llama 3.1-70B  & .591 & .569 & .750 & .458 & .724 & .178 & 39.3 \\
                & Qwen 3-32B     & .590 & .517 & .771 & .389 & .790 & .184 & 32.5 \\
                & UltraMed-70B   & .558 & .429 & .733 & .303 & .812 & .127 & 26.1 \\
                & Llama 3.1-8B   & .550 & .734 & .674 & .805 & .295 & .114 & 76.9 \\
                & MedGemma 4B    & .548 & .731 & .673 & .800 & .295 & .108 & 76.6 \\
                & Qwen 3-4B      & .539 & .717 & .668 & .774 & .305 & .086 & 74.6 \\
                & Qwen 3-8B      & .538 & .580 & .681 & .505 & .571 & .074 & 47.8 \\
                & Llama 3.2-3B   & .519 & .767 & .657 & .921 & .117 & .061 & 90.8 \\
                & Gemma 3-27B    & .518 & .752 & .654 & .884 & .152 & .052 & 87.1 \\
                & Phi-4-mini     & .512 & .741 & .649 & .862 & .162 & .033 & 85.4 \\
                & Gemma 3-4B     & .511 & .767 & .650 & .937 & .086 & .042 & 92.9 \\
                & MediPhi 4B     & .501 & .744 & .645 & .879 & .124 & .004 & 87.8 \\
                & UltraMed-8B    & .455 & .378 & .580 & .280 & .630 & $-$.093 & 31.2 \\
                & Mistral 7B     & .447 & .675 & .614 & .749 & .146 & $-$.123 & 78.6 \\
\end{longtable}

\begin{longtable}{llrrrrrrr}
\caption{Traditional metrics --- Neutral-Greedy configuration (all 15 models $\times$ 4 datasets = 60 runs). Sorted by balanced accuracy within each dataset. Flag\%(error-flag rate) is the fraction of notes flagged as containing an error.}\label{tab:trad240:ng} \\
\toprule
Dataset & Model & BalAcc & F1 & Prec & Rec & Spec & MCC & Flag\% \\
\midrule
\endfirsthead
\multicolumn{9}{l}{\textit{(Table \thetable\ continued from previous page)}} \\
\toprule
Dataset & Model & BalAcc & F1 & Prec & Rec & Spec & MCC & Flag\% \\
\midrule
\endhead
\midrule
\multicolumn{9}{r}{\textit{(continued on next page)}} \\
\endfoot
\bottomrule
\endlastfoot
MS-Test         & UltraMed-70B   & .642 & .650 & .661 & .640 & .643 & .283 & 50.4 \\
                & MedGemma 27B   & .596 & .690 & .585 & .839 & .353 & .221 & 74.7 \\
                & Llama 3.1-70B  & .586 & .669 & .584 & .784 & .389 & .188 & 70.2 \\
                & Qwen 3-32B     & .561 & .676 & .559 & .855 & .266 & .150 & 79.7 \\
                & UltraMed-8B    & .541 & .618 & .552 & .701 & .381 & .087 & 66.2 \\
                & Qwen 3-8B      & .533 & .672 & .540 & .891 & .175 & .094 & 85.9 \\
                & MedGemma 4B    & .517 & .656 & .531 & .859 & .175 & .046 & 84.3 \\
                & Gemma 3-4B     & .511 & .689 & .526 & .997 & .024 & .092 & 98.7 \\
                & Gemma 3-27B    & .510 & .675 & .526 & .942 & .077 & .038 & 93.3 \\
                & Qwen 3-4B      & .509 & .677 & .526 & .949 & .070 & .039 & 94.0 \\
                & Phi-4-mini     & .507 & .676 & .525 & .949 & .066 & .032 & 94.1 \\
                & Llama 3.1-8B   & .507 & .677 & .524 & .958 & .056 & .033 & 95.1 \\
                & Llama 3.2-3B   & .506 & .685 & .524 & .990 & .021 & .046 & 98.5 \\
                & Mistral 7B     & .501 & .682 & .521 & .987 & .014 & .005 & 98.7 \\
                & MediPhi 4B     & .495 & .670 & .519 & .945 & .045 & $-$.021 & 95.0 \\
\midrule
MedErrBench-EN  & UltraMed-70B   & .712 & .774 & .636 & .990 & .433 & .510 & 77.9 \\
                & UltraMed-8B    & .609 & .688 & .575 & .854 & .363 & .249 & 74.6 \\
                & Qwen 3-32B     & .596 & .706 & .555 & .971 & .221 & .291 & 87.5 \\
                & MedGemma 27B   & .582 & .697 & .546 & .962 & .202 & .251 & 88.0 \\
                & Llama 3.1-70B  & .545 & .691 & .528 & 1.000 & .089 & .217 & 95.6 \\
                & Qwen 3-8B      & .538 & .678 & .521 & .971 & .106 & .154 & 93.3 \\
                & Qwen 3-4B      & .534 & .680 & .518 & .990 & .077 & .165 & 95.7 \\
                & Phi-4-mini     & .529 & .678 & .515 & .990 & .067 & .150 & 96.2 \\
                & MediPhi 4B     & .524 & .675 & .512 & .990 & .058 & .133 & 96.6 \\
                & Llama 3.1-8B   & .515 & .673 & .507 & 1.000 & .029 & .122 & 98.5 \\
                & MedGemma 4B    & .510 & .643 & .505 & .885 & .135 & .029 & 87.5 \\
                & Mistral 7B     & .510 & .671 & .505 & 1.000 & .019 & .099 & 99.0 \\
                & Gemma 3-27B    & .510 & .671 & .505 & 1.000 & .019 & .099 & 99.0 \\
                & Gemma 3-4B     & .500 & .667 & .500 & 1.000 & .000 & .000 & 100.0 \\
                & Llama 3.2-3B   & .481 & .633 & .489 & .894 & .067 & $-$.068 & 91.3 \\
\midrule
MedErrBench-CN  & Qwen 3-32B     & .655 & .679 & .635 & .730 & .580 & .314 & 57.5 \\
                & UltraMed-70B   & .635 & .535 & .737 & .420 & .850 & .299 & 28.5 \\
                & Qwen 3-4B      & .590 & .664 & .562 & .810 & .370 & .200 & 72.0 \\
                & MedGemma 27B   & .590 & .677 & .558 & .860 & .320 & .214 & 77.0 \\
                & UltraMed-8B    & .549 & .570 & .549 & .592 & .507 & .099 & 54.3 \\
                & Llama 3.1-70B  & .532 & .559 & .532 & .590 & .475 & .065 & 55.8 \\
                & Llama 3.2-3B   & .530 & .654 & .512 & .904 & .156 & .091 & 87.4 \\
                & Qwen 3-8B      & .530 & .605 & .522 & .720 & .340 & .065 & 69.0 \\
                & MedGemma 4B    & .529 & .633 & .516 & .818 & .240 & .071 & 78.9 \\
                & Gemma 3-27B    & .520 & .657 & .511 & .920 & .120 & .067 & 90.0 \\
                & MediPhi 4B     & .520 & .669 & .511 & .970 & .070 & .092 & 95.0 \\
                & Llama 3.1-8B   & .510 & .585 & .507 & .690 & .330 & .021 & 68.0 \\
                & Gemma 3-4B     & .505 & .669 & .503 & 1.000 & .010 & .071 & 99.5 \\
                & Mistral 7B     & .490 & .658 & .497 & .970 & .010 & $-$.071 & 98.0 \\
                & Phi-4-mini     & .485 & .639 & .492 & .910 & .060 & $-$.057 & 92.5 \\
\midrule
MedRECT-JA      & Qwen 3-4B      & .600 & .801 & .696 & .942 & .257 & .285 & 87.1 \\
                & MedGemma 27B   & .596 & .757 & .703 & .821 & .371 & .214 & 75.3 \\
                & Qwen 3-32B     & .596 & .771 & .700 & .858 & .333 & .225 & 79.0 \\
                & Llama 3.1-70B  & .567 & .762 & .679 & .867 & .267 & .167 & 81.9 \\
                & UltraMed-70B   & .551 & .504 & .710 & .390 & .713 & .103 & 35.3 \\
                & Phi-4-mini     & .540 & .722 & .668 & .784 & .295 & .089 & 75.6 \\
                & UltraMed-8B    & .531 & .391 & .704 & .270 & .792 & .069 & 24.8 \\
                & Qwen 3-8B      & .528 & .771 & .658 & .932 & .124 & .094 & 91.2 \\
                & Mistral 7B     & .524 & .652 & .661 & .644 & .404 & .047 & 62.7 \\
                & Gemma 3-27B    & .521 & .786 & .654 & .984 & .057 & .115 & 96.9 \\
                & MediPhi 4B     & .514 & .750 & .651 & .884 & .143 & .039 & 87.5 \\
                & Gemma 3-4B     & .511 & .779 & .649 & .974 & .048 & .056 & 96.6 \\
                & Llama 3.2-3B   & .502 & .747 & .645 & .887 & .117 & .005 & 88.6 \\
                & Llama 3.1-8B   & .494 & .755 & .645 & .910 & .078 & $-$.020 & 91.4 \\
                & MedGemma 4B    & .489 & .720 & .638 & .826 & .152 & $-$.027 & 83.4 \\
\end{longtable}

\begin{longtable}{llrrrrrrr}
\caption{Traditional metrics --- Neutral-Sampling configuration (all 15 models $\times$ 4 datasets = 60 runs). Sorted by balanced accuracy within each dataset. Flag\%(error-flag rate) is the fraction of notes flagged as containing an error.}\label{tab:trad240:ns} \\
\toprule
Dataset & Model & BalAcc & F1 & Prec & Rec & Spec & MCC & Flag\% \\
\midrule
\endfirsthead
\multicolumn{9}{l}{\textit{(Table \thetable\ continued from previous page)}} \\
\toprule
Dataset & Model & BalAcc & F1 & Prec & Rec & Spec & MCC & Flag\% \\
\midrule
\endhead
\midrule
\multicolumn{9}{r}{\textit{(continued on next page)}} \\
\endfoot
\bottomrule
\endlastfoot
MS-Test         & UltraMed-70B   & .621 & .688 & .611 & .788 & .455 & .258 & 67.2 \\
                & MedGemma 27B   & .568 & .666 & .567 & .807 & .329 & .155 & 74.2 \\
                & UltraMed-8B    & .557 & .586 & .577 & .597 & .517 & .114 & 54.2 \\
                & Qwen 3-32B     & .545 & .666 & .549 & .846 & .245 & .113 & 80.2 \\
                & Llama 3.1-70B  & .543 & .675 & .548 & .878 & .208 & .116 & 83.7 \\
                & Qwen 3-8B      & .538 & .674 & .543 & .887 & .189 & .107 & 85.1 \\
                & Phi-4-mini     & .514 & .670 & .529 & .913 & .115 & .047 & 89.9 \\
                & Gemma 3-27B    & .511 & .676 & .527 & .942 & .080 & .045 & 93.1 \\
                & MediPhi 4B     & .511 & .669 & .527 & .916 & .105 & .036 & 90.6 \\
                & Llama 3.2-3B   & .511 & .669 & .527 & .916 & .105 & .036 & 90.6 \\
                & Qwen 3-4B      & .507 & .676 & .525 & .949 & .066 & .032 & 94.1 \\
                & Llama 3.1-8B   & .507 & .667 & .525 & .913 & .102 & .025 & 90.6 \\
                & Mistral 7B     & .504 & .686 & .523 & .997 & .010 & .045 & 99.3 \\
                & Gemma 3-4B     & .500 & .683 & .521 & .990 & .010 & .004 & 99.0 \\
                & MedGemma 4B    & .497 & .639 & .519 & .830 & .164 & $-$.008 & 83.2 \\
\midrule
MedErrBench-EN  & UltraMed-70B   & .692 & .759 & .623 & .971 & .413 & .463 & 77.9 \\
                & UltraMed-8B    & .615 & .667 & .575 & .794 & .436 & .245 & 67.7 \\
                & MedGemma 27B   & .611 & .716 & .564 & .981 & .240 & .329 & 87.0 \\
                & Qwen 3-32B     & .587 & .703 & .548 & .981 & .192 & .281 & 89.4 \\
                & Llama 3.1-70B  & .567 & .696 & .536 & .990 & .144 & .253 & 92.3 \\
                & Qwen 3-8B      & .553 & .683 & .529 & .962 & .144 & .184 & 90.9 \\
                & Phi-4-mini     & .529 & .671 & .515 & .962 & .096 & .115 & 93.3 \\
                & Qwen 3-4B      & .524 & .675 & .512 & .990 & .058 & .133 & 96.6 \\
                & Llama 3.1-8B   & .514 & .667 & .508 & .971 & .058 & .071 & 95.7 \\
                & MedGemma 4B    & .505 & .644 & .503 & .894 & .115 & .015 & 88.9 \\
                & MediPhi 4B     & .505 & .656 & .503 & .942 & .067 & .020 & 93.8 \\
                & Gemma 3-27B    & .505 & .669 & .502 & 1.000 & .010 & .070 & 99.5 \\
                & Gemma 3-4B     & .500 & .667 & .500 & 1.000 & .000 & .000 & 100.0 \\
                & Mistral 7B     & .500 & .665 & .500 & .990 & .010 & .000 & 99.0 \\
                & Llama 3.2-3B   & .475 & .589 & .481 & .757 & .192 & $-$.061 & 78.3 \\
\midrule
MedErrBench-CN  & Qwen 3-32B     & .655 & .682 & .632 & .740 & .570 & .315 & 58.5 \\
                & Qwen 3-4B      & .605 & .675 & .573 & .820 & .390 & .233 & 71.5 \\
                & UltraMed-70B   & .587 & .560 & .593 & .531 & .643 & .175 & 44.3 \\
                & MedGemma 27B   & .567 & .656 & .547 & .820 & .313 & .155 & 75.4 \\
                & Llama 3.1-8B   & .556 & .627 & .540 & .747 & .364 & .120 & 69.2 \\
                & Llama 3.1-70B  & .550 & .637 & .534 & .790 & .310 & .114 & 74.0 \\
                & MedGemma 4B    & .545 & .640 & .529 & .810 & .280 & .106 & 76.5 \\
                & UltraMed-8B    & .537 & .500 & .553 & .457 & .618 & .075 & 42.0 \\
                & Qwen 3-8B      & .525 & .592 & .519 & .690 & .360 & .053 & 66.5 \\
                & Gemma 3-27B    & .515 & .657 & .508 & .930 & .100 & .054 & 91.5 \\
                & Phi-4-mini     & .515 & .664 & .508 & .960 & .070 & .066 & 94.5 \\
                & MediPhi 4B     & .510 & .645 & .506 & .890 & .130 & .031 & 88.0 \\
                & Gemma 3-4B     & .505 & .669 & .503 & 1.000 & .010 & .071 & 99.5 \\
                & Mistral 7B     & .500 & .664 & .500 & .990 & .010 & .000 & 99.0 \\
                & Llama 3.2-3B   & .490 & .593 & .467 & .814 & .167 & $-$.025 & 82.4 \\
\midrule
MedRECT-JA      & Qwen 3-32B     & .595 & .767 & .700 & .847 & .343 & .220 & 78.0 \\
                & MedGemma 27B   & .573 & .734 & .690 & .784 & .362 & .158 & 73.2 \\
                & Qwen 3-4B      & .556 & .761 & .675 & .874 & .238 & .144 & 83.4 \\
                & UltraMed-70B   & .551 & .640 & .683 & .602 & .500 & .099 & 56.6 \\
                & Llama 3.1-70B  & .538 & .776 & .665 & .932 & .144 & .123 & 90.5 \\
                & Qwen 3-8B      & .538 & .778 & .663 & .942 & .133 & .130 & 91.5 \\
                & Phi-4-mini     & .534 & .730 & .664 & .811 & .257 & .079 & 78.6 \\
                & Llama 3.2-3B   & .533 & .740 & .668 & .829 & .238 & .080 & 80.6 \\
                & MedGemma 4B    & .521 & .754 & .655 & .889 & .152 & .061 & 87.5 \\
                & Gemma 3-27B    & .520 & .782 & .654 & .974 & .067 & .098 & 95.9 \\
                & Llama 3.1-8B   & .513 & .763 & .651 & .921 & .105 & .044 & 91.2 \\
                & MediPhi 4B     & .511 & .726 & .650 & .821 & .200 & .026 & 81.4 \\
                & Mistral 7B     & .500 & .663 & .645 & .683 & .317 & $-$.000 & 68.3 \\
                & Gemma 3-4B     & .497 & .766 & .643 & .947 & .048 & $-$.011 & 94.9 \\
                & UltraMed-8B    & .489 & .312 & .621 & .208 & .771 & $-$.025 & 21.6 \\
\end{longtable}

}

\FloatBarrier

\subsection{Full \eca{} Category Breakdown}
\label{app:eca:full} 

Table~\ref{tab:eca:full} reports the complete \eca{} breakdown for all 15 models on all four datasets, with category counts pooled across the four prompt--decoding configurations. TP localization and FP evidence-hit are reported as four-configuration means, matching the convention used in Section~\ref{sec:results:eca}; per-configuration raw outputs are available as supplementary material. 

{\footnotesize
\setlength{\tabcolsep}{3pt}
\renewcommand{\arraystretch}{0.95}

\begin{longtable}{llrrrrrrrr}
\caption{Full \eca{} breakdown per model--dataset combination (totals pooled across the four configurations). $n$: Pred1 pairs categorized into A--E. TP~loc and FP~hit: four-configuration mean overlap rates (\%). A--E: pooled category counts (Both-Hit~/~TP-Only~/~FP-Only~/~Neither-Hit~/~Extraction-Fail). Within each dataset, rows are sorted by TP~loc. Every model--dataset combination had at least one configuration with $\ge 5$ \textit{Pred1} pairs, so all 60 combinations are listed. Because TP~loc is an unweighted four-configuration mean while A--E are pooled counts, TP~loc and $(A+B)/n$ can differ by several points when \textit{Pred1} counts are uneven across configurations.}
\label{tab:eca:full} \\
\toprule
Dataset & Model & $n$ & TP loc & FP hit & A & B & C & D & E \\
\midrule
\endfirsthead
\multicolumn{10}{l}{\textit{(Table \thetable\ continued from previous page)}} \\
\toprule
Dataset & Model & $n$ & TP loc & FP hit & A & B & C & D & E \\
\midrule
\endhead
\midrule
\multicolumn{10}{r}{\textit{(continued on next page)}} \\
\endfoot
\bottomrule
\endlastfoot
MS-Test         & Qwen 3-32B     &  521 &  86.9 &  48.8 &  240 &  213 &   11 &   46 &   11 \\
                & Qwen 3-4B      &  818 &  81.4 &  49.9 &  383 &  286 &   24 &  101 &   24 \\
                & Qwen 3-8B      &  452 &  78.8 &  41.5 &  202 &  182 &   19 &   43 &    6 \\
                & Gemma 3-27B    &  668 &  75.7 &  43.7 &  273 &  221 &   23 &  133 &   18 \\
                & Gemma 3-4B     & 1033 &  75.2 &  58.0 &  507 &  269 &   90 &  107 &   60 \\
                & MedGemma 27B   &  311 &  73.3 &  42.9 &  145 &   92 &   11 &   53 &   10 \\
                & Phi-4-mini     &  761 &  72.3 &  45.4 &  289 &  260 &   60 &  107 &   45 \\
                & Mistral 7B     & 1027 &  71.0 &  50.7 &  436 &  292 &   82 &  157 &   60 \\
                & MedGemma 4B    &  524 &  68.0 &  37.8 &  178 &  180 &   19 &  119 &   28 \\
                & MediPhi 4B     &  830 &  63.8 &  45.5 &  311 &  218 &   63 &  146 &   92 \\
                & Llama 3.1-70B  &  397 &  62.9 &  36.3 &  126 &  143 &   16 &   92 &   20 \\
                & UltraMed-8B    &  250 &  62.5 &  30.3 &   71 &   90 &   12 &   49 &   28 \\
                & UltraMed-70B   &  213 &  59.8 &  35.4 &   65 &   72 &   11 &   53 &   12 \\
                & Llama 3.1-8B   &  616 &  56.0 &  42.4 &  196 &  141 &   51 &  143 &   85 \\
                & Llama 3.2-3B   & 1007 &  46.0 &  34.0 &  257 &  208 &   85 &  273 &  184 \\
\midrule
MedErrBench-EN  & Qwen 3-32B     &  255 &  96.8 &  90.2 &  224 &   22 &    5 &    4 &    0 \\
                & Qwen 3-8B      &  206 &  95.6 &  95.3 &  188 &    7 &    6 &    5 &    0 \\
                & UltraMed-70B   &  145 &  95.1 &  85.8 &  112 &   26 &    7 &    0 &    0 \\
                & MedGemma 27B   &  162 &  94.8 &  86.2 &  126 &   20 &    8 &    7 &    1 \\
                & Qwen 3-4B      &  345 &  94.0 &  88.3 &  295 &   30 &   10 &    8 &    2 \\
                & Gemma 3-27B    &  356 &  92.8 &  81.7 &  277 &   52 &   11 &   14 &    2 \\
                & Gemma 3-4B     &  398 &  90.8 &  89.2 &  330 &   31 &   25 &    8 &    4 \\
                & Llama 3.1-70B  &  256 &  87.4 &  79.4 &  185 &   38 &   15 &   16 &    2 \\
                & Phi-4-mini     &  334 &  81.9 &  78.8 &  221 &   53 &   42 &   12 &    6 \\
                & MediPhi 4B     &  334 &  81.5 &  82.1 &  236 &   37 &   39 &   13 &    9 \\
                & Llama 3.2-3B   &  326 &  80.7 &  82.3 &  231 &   34 &   38 &   10 &   13 \\
                & Mistral 7B     &  382 &  80.4 &  76.0 &  259 &   47 &   30 &   38 &    8 \\
                & UltraMed-8B    &  114 &  80.3 &  66.8 &   68 &   25 &   10 &    6 &    5 \\
                & MedGemma 4B    &  215 &  80.0 &  79.4 &  146 &   24 &   28 &   16 &    1 \\
                & Llama 3.1-8B   &  271 &  79.7 &  69.6 &  162 &   49 &   20 &   29 &   11 \\
\midrule
MedErrBench-CN  & Qwen 3-4B      &  185 &  93.0 &  80.6 &  144 &   28 &    4 &    9 &    0 \\
                & UltraMed-70B   &   36 &  92.1 &  86.3 &   22 &    8 &    2 &    4 &    0 \\
                & Qwen 3-32B     &   94 &  90.3 &  74.9 &   69 &   15 &    2 &    8 &    0 \\
                & Phi-4-mini     &  322 &  86.2 &  87.8 &  262 &   16 &   20 &   24 &    0 \\
                & Gemma 3-4B     &  340 &  82.2 &  79.1 &  242 &   35 &   26 &   34 &    3 \\
                & Llama 3.1-8B   &  124 &  75.9 &  71.8 &   69 &   15 &   13 &   19 &    8 \\
                & MedGemma 4B    &  153 &  70.1 &  63.2 &   92 &   17 &    8 &   35 &    1 \\
                & Gemma 3-27B    &  272 &  69.5 &  65.0 &  147 &   41 &   24 &   57 &    3 \\
                & MedGemma 27B   &  134 &  67.8 &  59.8 &   80 &   22 &    2 &   26 &    4 \\
                & Llama 3.1-70B  &  120 &  63.8 &  49.9 &   71 &   28 &    1 &   19 &    1 \\
                & Mistral 7B     &  345 &  59.4 &  57.9 &  169 &   37 &   30 &   82 &   27 \\
                & Qwen 3-8B      &  122 &  54.9 &  50.3 &   84 &   18 &    7 &   11 &    2 \\
                & MediPhi 4B     &  301 &  51.1 &  51.9 &  127 &   30 &   31 &   91 &   22 \\
                & Llama 3.2-3B   &  171 &  49.2 &  57.3 &   79 &   17 &   27 &   31 &   17 \\
                & UltraMed-8B    &   35 &  42.5 &  29.4 &    9 &   11 &    3 &   11 &    1 \\
\midrule
MedRECT-JA      & Qwen 3-32B     &  270 &  91.8 &  44.8 &  115 &  127 &    5 &   14 &    9 \\
                & Qwen 3-4B      &  493 &  68.7 &  41.7 &  188 &  150 &   18 &  121 &   16 \\
                & Qwen 3-8B      &  423 &  64.7 &  45.5 &  139 &  130 &   34 &   95 &   25 \\
                & Llama 3.1-70B  &  335 &  64.6 &  33.8 &   69 &  108 &   15 &  121 &   22 \\
                & MedGemma 27B   &  323 &  58.7 &  32.0 &   84 &  110 &   17 &   93 &   19 \\
                & Gemma 3-27B    &  656 &  53.8 &  22.2 &  119 &  231 &   25 &  220 &   61 \\
                & Gemma 3-4B     &  666 &  40.0 &  23.9 &  102 &  163 &   56 &  271 &   74 \\
                & Phi-4-mini     &  500 &  35.1 &  24.5 &   67 &  109 &   56 &  164 &  104 \\
                & Llama 3.1-8B   &  549 &  32.1 &  28.5 &   85 &   85 &   66 &  183 &  130 \\
                & Llama 3.2-3B   &  588 &  25.1 &  17.1 &   75 &   75 &   31 &  265 &  142 \\
                & MedGemma 4B    &  528 &  23.0 &  18.0 &   46 &   70 &   45 &  276 &   91 \\
                & UltraMed-70B   &  106 &  21.1 &  16.8 &   10 &   17 &    8 &   48 &   23 \\
                & UltraMed-8B    &   59 &  20.1 &   8.4 &    2 &    9 &    5 &   28 &   15 \\
                & Mistral 7B     &  431 &  19.5 &  24.2 &   44 &   41 &   62 &  190 &   94 \\
                & MediPhi 4B     &  586 &  18.3 &  13.7 &   35 &   72 &   44 &  287 &  148 \\
\end{longtable}
}

\end{document}